\pdfoutput=1
\documentclass[10pt,twocolumn,letterpaper]{article}

\usepackage[pagenumbers]{wacv}

\usepackage{multirow}        %
\usepackage{colortbl}        %
\usepackage{microtype}       %
\usepackage[utf8]{inputenc}  %
\usepackage[T1]{fontenc}     %
\usepackage{xr-hyper}        %

\newcommand{\fitcol}[1]{\resizebox{\ifdim\width>\linewidth\linewidth\else\width\fi}{!}{#1}}

\newcommand{\PublishedRtfmUcfIthreed}{84.30}
\newcommand{\PublishedMgfnUcfIthreed}{86.98}
\newcommand{\PublishedMgfnXdvIthreed}{79.19}

\newcommand{\DispRLtwo}{1.0002}
\newcommand{\DispRCosine}{1.0014}

\newcommand{\CtrlUncondFloor}{0.15}
\newcommand{\CtrlRandUntrainedLo}{1.75}
\newcommand{\CtrlRandUntrainedHi}{14.92}
\newcommand{\CtrlRandTrained}{0.09}

\newcommand{\RandkRelLo}{1.01}
\newcommand{\RandkRelHi}{16.50}

\newcommand{\FidelityGateTthree}{4.23}

\newcommand{\FidelityKoneCap}{0.28}
\newcommand{\FidelityKoneOffCap}{1.80}

\newcommand{\RandCtrlNTotal}{290}

\newcommand{\RandCtrlRelLo}{4.23}
\newcommand{\RandCtrlRelHi}{6.78}
\newcommand{\RandCtrlRandLo}{-1.45}
\newcommand{\RandCtrlRandHi}{1.13}

\newcommand{\CleanAucSpanUcfClip}{5.78}

\newcommand{\TeacherEdgeFoneLo}{0.138}
\newcommand{\TeacherEdgeFoneHi}{0.250}
\newcommand{\TeacherEdgeNullLo}{0.111}
\newcommand{\TeacherEdgeNullHi}{0.167}
\newcommand{\TeacherPairedWinLo}{73.4}
\newcommand{\TeacherPairedWinHi}{81.4}

\newcommand{\MeanRootsPerVideo}{3.74}
\newcommand{\MeanIntermediatesPerVideo}{4.71}
\newcommand{\MeanLeavesPerVideo}{3.80}
\newcommand{\EdgeSlotPairs}{1250}

\newcommand{\EdgeSparsityPct}{0.7}

\newcommand{\SpreadFidelityTthree}{1.6}

\newcommand{\PublishedRtfmXdvIthreed}{77.81}
\newcommand{\ShortfallMgfnUcfIthreed}{6.2}
\newcommand{\ShortfallRtfmUcfIthreed}{4.6}
\newcommand{\ShortfallMgfnXdvIthreed}{11.6}
\newcommand{\ShortfallRtfmXdvIthreed}{3.8}
\newcommand{\FidelityIrrCap}{0.09}

\newcommand{\QthirtyfourDpos}{1}
\newcommand{\QthirtyfourDneg}{2}
\newcommand{\QthirtyfourDzero}{3}

\newcommand{\LatCosRelAnom}{-0.331}
\newcommand{\LatCosRandAnom}{-0.307}

\newcommand{\LatLtwoRelAnom}{14.3}
\newcommand{\LatLtwoRandAnom}{9.8}
\newcommand{\LatCosAnomGap}{0.024}
\newcommand{\LatRandFartherPct}{68}

\newcommand{\VadClipAucHarness}{85.82}
\newcommand{\BootReplicates}{10{,}000}

\newcommand{\HolmFamilyAlpha}{0.05}
\newcommand{\BootPercentileLo}{2.5}
\newcommand{\BootPercentileHi}{97.5}
\newcommand{\BootRatioGuardMax}{13}

\newcommand{\SplitNormalShiftRatioLo}{2.4}
\newcommand{\SplitNormalShiftRatioHi}{5.7}

\newcommand{\FidelityRegenLo}{-0.10}
\newcommand{\FidelityRegenHi}{0.30}

\newcommand{\DivFreeRelMgfn}{0.48}
\newcommand{\DivFreeRelUrdmu}{0.23}
\newcommand{\DivFreeRelRtfm}{2.22}
\newcommand{\DivFreeRelDsanet}{1.65}
\newcommand{\DivFreeRelVadclip}{1.06}
\newcommand{\DivFreeRelGlancevad}{1.26}
\newcommand{\DivFreeRandMgfn}{0.56}
\newcommand{\DivFreeRandUrdmu}{0.30}
\newcommand{\DivFreeRandRtfm}{0.57}
\newcommand{\DivFreeRandDsanet}{1.00}
\newcommand{\DivFreeRandVadclip}{0.82}
\newcommand{\DivFreeRandGlancevad}{0.35}

\newcommand{\DivFreeEncCos}{0.8817}

\newcommand{\DivFreeDiffMgfn}{-0.09}
\newcommand{\DivFreeDiffUrdmu}{-0.06}
\newcommand{\DivFreeDiffRtfm}{1.64}
\newcommand{\DivFreeDiffDsanet}{0.65}
\newcommand{\DivFreeDiffVadclip}{0.23}
\newcommand{\DivFreeDiffGlancevad}{0.91}
\newcommand{\DivFreeCiLoMgfn}{-2.63}
\newcommand{\DivFreeCiLoUrdmu}{-1.89}
\newcommand{\DivFreeCiLoRtfm}{-1.08}
\newcommand{\DivFreeCiLoDsanet}{-1.61}
\newcommand{\DivFreeCiLoVadclip}{-2.22}
\newcommand{\DivFreeCiLoGlancevad}{-0.39}
\newcommand{\DivFreeCiHiMgfn}{2.31}
\newcommand{\DivFreeCiHiUrdmu}{1.64}
\newcommand{\DivFreeCiHiRtfm}{4.30}
\newcommand{\DivFreeCiHiDsanet}{2.70}
\newcommand{\DivFreeCiHiVadclip}{2.85}
\newcommand{\DivFreeCiHiGlancevad}{2.18}

\newcommand{\ReconRelMaxTthree}{6.78}



\usepackage{tcolorbox}
\newtcolorbox{promptbox}[1]{colback=gray!4, colframe=gray!55, colbacktitle=gray!18,
  coltitle=black, fonttitle=\bfseries\small, title=#1, boxrule=0.4pt, arc=1pt,
  left=4pt, right=4pt, top=2pt, bottom=2pt, before skip=3pt, after skip=3pt,
  fontupper=\small}
\newcommand{\promptout}[1]{\par\smallskip\noindent\textit{Output:} \texttt{\footnotesize #1}}

\definecolor{wacvblue}{rgb}{0.21,0.49,0.74}
\usepackage[pagebackref,breaklinks,colorlinks,allcolors=wacvblue]{hyperref}

\def\wacvPaperID{14}
\def\confName{WACV}
\def\confYear{2027}

\title{How to Explain your Unexplainable Video Anomaly Detector}

\author{Giacomo D'Amicantonio \qquad Egor Bondarev\\
Eindhoven University of Technology\\
Eindhoven, The Netherlands\\
{\tt\small g.d.amicantonio@tue.nl}
}

\begin{document}
\maketitle
\begin{abstract}
Weakly-supervised Video Anomaly Detection (VAD) models assign anomaly scores from high-dimensional features, and their predictions remain unexplained. Saliency maps show where a model looks but not what it relies on, and the descriptions produced by Vision-Language Models (VLMs) are generated by branches that the anomaly score never reads. To address this faithfulness gap, we propose \textbf{Explain The Unexplainable (ETU)}, a classifier-agnostic framework that audits frozen VAD models through counterfactual interventions in their feature space. ETU maps video features to a discrete dependency graph of agents, actions and outcomes ($A_r \to A_a \to A_o$), regenerates the features by graph-conditioned diffusion with selected nodes of the graph masked, and measures the change in the score of a frozen detector; one set of interventions is applied unchanged to every detector. The generator is trained only to reconstruct the input features, so it cannot learn the contrast it is asked to measure, and we validate the audit with experiments that test every alternative explanation of its results, from the cost of regeneration to random deletions of matched size. On UCF-Crime features obtained from a CLIP encoder, we show that substituting the scene semantics results in the largest drop for most detectors, and that removing the outcome chain of the graph ($\Delta_{rel}$) changes the anomaly score of all six audited detectors far more than removing the same number of random nodes, even though the random deletion perturbs the features about as much. We show that this targeted behavior comes from the divergence objective of the graph embedding. Without it, the two interventions become indistinguishable, and with a divergence objective in the generator as well, the score change of the detector follows the displacement of the reconstruction rather than the semantics that were removed.
\end{abstract}

\section{Introduction}
\label{sec:intro}
Video Anomaly Detection (VAD) detects events that are unusual within a video's context; because anomalies are sparse and unpredictable, its public datasets are mostly real-world CCTV footage of crimes and accidents. The dominant paradigm, Weakly-Supervised VAD (WSVAD), tackles this task relying solely on video-level labels. This paradigm is based on Multiple Instance Learning (MIL), pushing a WSVAD model to assign high anomaly scores to the most abnormal snippets of a known anomalous video and low scores for snippets of known normal videos. However, a critical drawback of MIL-based models is the inherently unexplainable nature of their predictions, which rely on high-dimensional feature spaces that are prone to spurious correlations.

Explainable VAD aims to untangle such correlations. Common approaches produce saliency maps over the pixels that most influence the score~\cite{wang2023unveiling, doshi2023towards}, answering ``\textit{where is the model looking?}''. But VAD models rarely process raw pixels; they consume high-dimensional, pre-trained encoders such as I3D~\cite{i3d} or CLIP~\cite{CLIP}. In this latent context, post-hoc attribution techniques such as Grad-CAM~\cite{selvaraju2017grad} are fundamentally limited, as they capture spatial correlation rather than the score's dependence on content. Liu and Ma~\cite{Fully_super_MM19} showed that VAD models exploit the scene background as a cue for the anomaly rather than the event itself, which suggests that they rely on spurious correlations in the feature space rather than on the actual events.

\begin{figure}[!t]
    \centering
    \includegraphics[width=\linewidth]{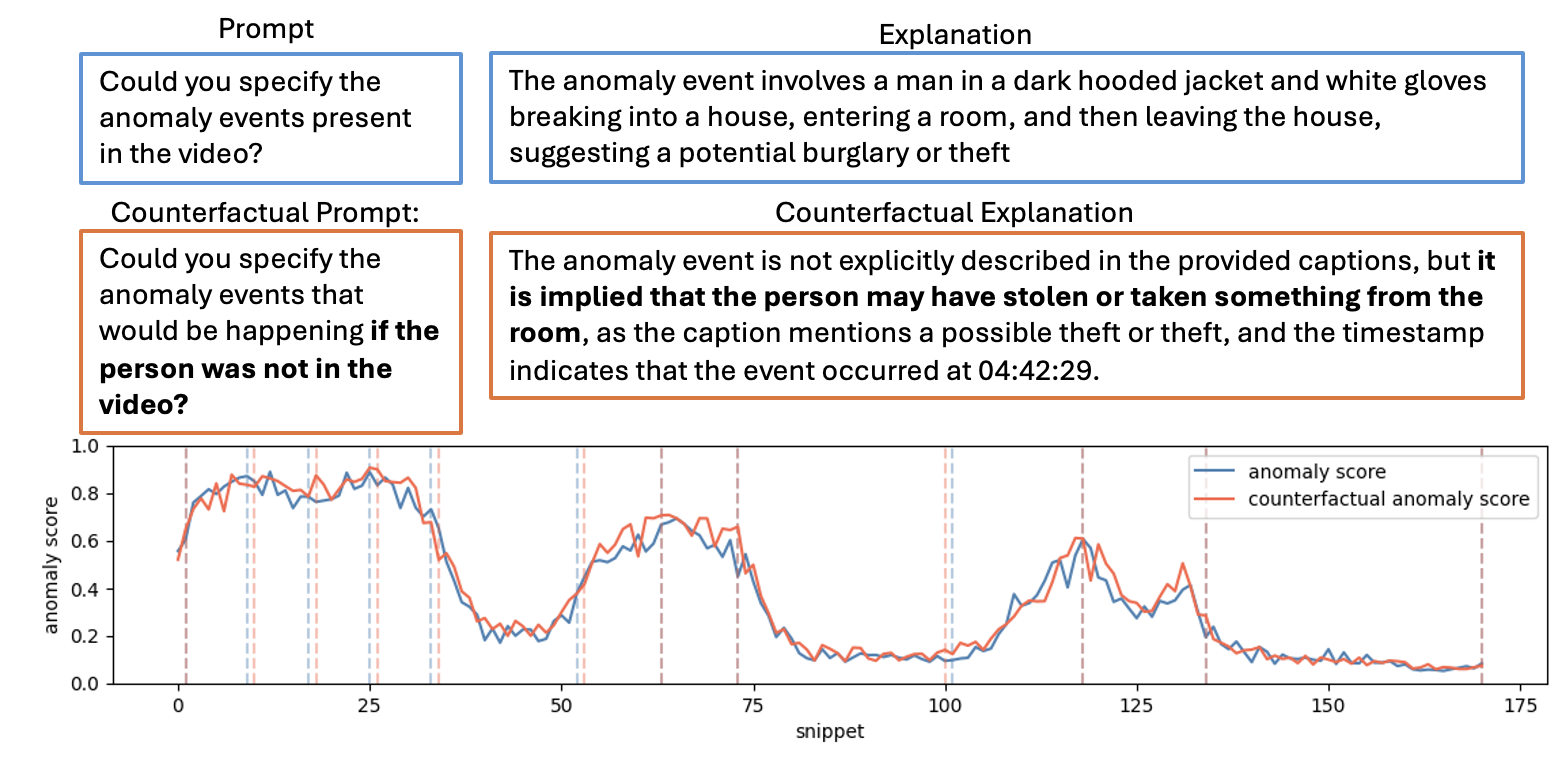}
    \caption{\textbf{A VLM explainer's caption changes under a counterfactual prompt but its score does not.} Counterfactual prompting of Holmes-VAU~\cite{zhang2025holmes} on the ``robbery'' video provided in the official repository. The anomaly scores obtained with the two prompts are nearly identical, so the score does not follow the explanation the model gives, while the counterfactual explanation mentions the very person the prompt instructed to ignore.}
    \label{fig:holmes1}
\end{figure}
Recently, the field has pivoted toward Vision-Language Models (VLMs) to provide descriptive grounding for anomaly detections~\cite{zhang2024holmes, zhang2025holmes}. While these methods address the \textit{"what"} of a scene, they typically employ parallel architectural branches that generate captions and anomaly scores through functionally decoupled pathways. This structural separation introduces a faithfulness gap. The generated caption serves as a post-hoc rationalization rather than an input the model's decision depends on, as shown in Figure~\ref{fig:holmes1}.

To bridge this faithfulness gap, we propose \textit{Explain The Unexplainable} (ETU, Figure~\ref{fig:etu_overview}), a classifier-agnostic framework designed to audit frozen VAD models by intervening directly in their latent feature space. ETU enforces a strict semantic bottleneck by mapping continuous video features into a discrete semantic manifold structured as a hierarchical dependency graph $\mathcal{G}$ representing compositional semantics, namely the agents involved (Roots, $A_r$), their specific behaviors (Actions, $A_a$), and the resulting consequences (Outcomes, $A_o$). To establish a structural prior for this space, we utilize an offline VLM that observes each video and emits valid $A_r \to A_a \to A_o$ chains directly. ETU then regenerates the features by graph-conditioned diffusion under a masked graph (e.g., with an anomaly-relevant sequence removed). The frozen model's response to these reconstructions moves the question from the spatial correlations of saliency maps (``\textit{where is the model looking?}'') and the decoupled descriptions of VLMs (``\textit{what is happening?}'') to intervention-based auditing (``\textit{how much did this score depend on the semantics the graph names?}'').

ETU is a two-stage pipeline. In the first stage, the Graph Embedding Model (GEM) maps video features into a discrete atom space and infers the directed dependency edges between roots, actions, and outcomes. The topology is aggregated into a compact latent embedding ($\mathbf{z}_{\text{graph}}$), distilling the offline VLM's structural priors into a lightweight model with no test-time text generation. In the second stage, a Graph-conditioned Feature Diffusion (GFD) model performs guided denoising of noisy ground-truth features conditioned on this embedding. It is trained only to reconstruct the encoder features under the complete graph. During inference, we condition the diffusion process on three deterministic masked variants of the graph, which remove the core anomaly chains ($\mathbf{z}_{\text{rel}}$), prune irrelevant background events ($\mathbf{z}_{\text{irr}}$), or replace the scene semantics ($\mathbf{z}_{\text{sem}}$). The downstream VAD model's response to these reconstructions is an anomaly delta ($\Delta$)~\cite{wu2021learning} that measures how much the prediction depended on the masked part of the conditioning graph.

An audit of this kind has a failure mode of its own, and we treat it as part of the method. A generator trained to make the relevant-mask reconstructions diverge can manufacture the very contrast it is asked to measure, and a frozen detector scored on regenerated features may react to the distribution shift rather than to the content~\cite{hooker2019roar,rong2022road}. ETU therefore keeps its generator out of that objective, training it on reconstruction alone, and every quantity we report is compared with a lower bound or with a test that can fail (Section~\ref{sec:controls}), including a generator that does carry a divergence objective and whose deltas follow the displacement of the reconstruction.

The main contributions of this paper are:
\begin{itemize}
    \setlength{\itemsep}{0pt}
    \setlength{\parskip}{0pt}
    \item \textbf{A framework.} ETU, a classifier-agnostic framework for the counterfactual auditing of frozen VAD models. A graph embedding model maps the video features to a discrete $A_r \to A_a \to A_o$ dependency graph, and a graph-conditioned diffusion model, trained only to reconstruct the features, regenerates them under masked graphs, so that one set of interventions is handed unchanged to every detector and read as a change of its score.
    \item \textbf{Validation.} The lower bounds, count-matched random deletions and their displacement, ablations of the divergence objective in the graph embedding and in the generator, and a comparison of the graphs of three teachers that a feature-space counterfactual audit needs before its numbers can be attributed to the detector, each stated with the alternative explanation it rules out and with its outcome.
    \item \textbf{A targeted intervention.} Substituting the scene semantics results in the largest drop for most detectors, and removing the outcome chain costs all six detectors more than removing the same number of random nodes, although the random deletion displaces the reconstructed features about as much. We show that the divergence objective of the graph embedding is the design choice that makes this possible. Without it the two deletions become indistinguishable, and a divergence objective in the generator makes the deltas follow the displacement instead. The delta is therefore read as the dependence of the score on graph-conditioned content, and not as the removal of a concept.
\end{itemize}

\section{Related Work}
\label{sec:related}

\subsection{Weakly-Supervised Video Anomaly Detection}
The dominant paradigm in Video Anomaly Detection (VAD) is Weakly-Supervised learning (WSVAD), which utilizes only video-level labels for training~\cite{sultani2018real, cvpr19, wu2020not, icip2019, bmvc19, iccv21_Dance, ICME2020, wu2021learning}. Sultani et al.~\cite{sultani2018real} established the foundation, using MIL to identify the most salient anomalous segments within video ``bags.'' Subsequent work refines the discriminative power of that objective with feature magnitude and discriminability~\cite{tian2021weakly, AAAI23MGFN}, unbiased classifiers against label noise~\cite{CVPR23UMIL}, self-training and mixture-of-experts designs~\cite{AAAI22MSL, CVPR23LAA, amicantonio2025mixture}, and multimodal features~\cite{majhi2025just, zhang2024holmes}. WSVAD models nevertheless map high-dimensional features directly to scalar scores. They localize in time but offer no semantic justification for a prediction.

\subsection{Explainable Video Anomaly Detection}
Explainable VAD has focused on spatial localization (``\textit{where}'') and semantic description (``\textit{what}''). Traditional frameworks interpret via pixel-level reconstruction errors or gradient-based saliency~\cite{szymanowicz2022discrete}, which capture spatial correlation rather than the score's dependence on semantic content. To inject semantics, attribute-based methods evaluate normality against pre-defined taxonomies of trajectories~\cite{singh2023eval}, poses~\cite{reiss2022attribute}, or interactions~\cite{szymanowicz2021x, hinami2017joint}, whose dataset-specific ontologies struggle to generalize to novel anomaly classes. Vision-Language Models (VLMs) instead generate rich, textual explanations~\cite{lv2024video, ye2025vera, yang2024follow, zanella2024harnessing, zhang2025holmes}, but usually exhibit the \textit{faithfulness gap} of Figure~\ref{fig:holmes1}. Captions and anomaly scores come from parallel, functionally decoupled branches, so the outputs are post-hoc rationalizations that describe the scene rather than audit the detector's internal logic. VERA~\cite{ye2025vera} is an exception, feeding the questions it learns into the score it reports.

\subsection{Faithfulness and Concept Interventions}
Deletion-based faithfulness tests score a model on inputs from which the attributed content has been removed. ROAR~\cite{hooker2019roar} showed that the resulting drop conflates importance with distribution shift, and retrains the model to separate the two. ROAD~\cite{rong2022road} showed that the shape of the deletion mask leaks class information, and replaced retraining with imputation. F-Fidelity~\cite{zheng2025ffidelity} and GOAR~\cite{park2023goar} harden the evaluation against the out-of-distribution inputs that deletion creates. ETU regenerates the input instead of retraining the model (Eq.~\ref{eq:delta}), and its count-matched random deletion plays the role of the leakage check of ROAD.

Concept-level methods such as TCAV~\cite{kim2018tcav}, ACE~\cite{ghorbani2019ace} and CRAFT~\cite{fel2023craft}, concept bottlenecks with their post-hoc, counterfactual and intervenable variants~\cite{koh2020concept,yuksekgonul2023posthoc,dominici2025counterfactualcbm,laguna2024intervenable}, and linear concept erasure~\cite{belrose2023leace} intervene on concepts inside the representation of the model itself, and their conclusions depend on the probe set~\cite{ramaswamy2023overlooked}. The quantity we estimate is the causal concept effect of Goyal et al.~\cite{goyal2019cace}, which ETU obtains from a regeneration instead of a generative in-fill such as FIDO~\cite{chang2019counterfactual}, whose off-distribution hazard Hase et al.~\cite{hase2021ood} document. Diffusion counterfactuals edit images or videos for a given classifier~\cite{jeanneret2022diffusion,jeanneret2023ace,weng2024fastdime,varshney2025ldvice,wang2026backtothefeature}, and counterfactual generation has been used to explain image anomaly detectors~\cite{ji2024arpro,liznerski2024reimagining}. Prior concept-intervention methods intervene inside the model being explained, so the intervention and the explanation share parameters. ETU instead intervenes on the input of a model it never trains against, which is what makes one set of interventions comparable across independently trained detectors. Its generator is trained to reconstruct the input and never to separate the interventions, and the objective that shapes its conditioning is a design choice that we verify in Section~\ref{sec:experiments}.

\begin{figure*}[!t]
    \centering
    \includegraphics[width=\textwidth]{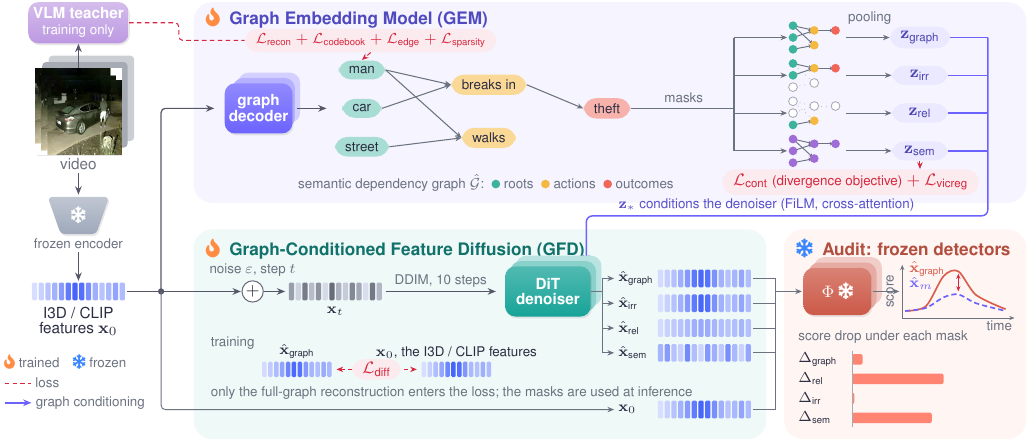}
\caption{\textbf{Explain The Unexplainable (ETU) hands one set of masked reconstructions per video unchanged to every frozen detector.} The Graph Embedding Model (GEM) maps continuous video features to a semantic dependency graph and a latent embedding $\mathbf{z}_{\text{graph}}$; deterministic masks yield the counterfactual embeddings $\mathbf{z}_{\text{rel}}, \mathbf{z}_{\text{irr}}, \mathbf{z}_{\text{sem}}$, on which the Graph-Conditioned Feature Diffusion (GFD) denoises the video features. GFD is trained only to reconstruct the encoder features under the complete graph; the masked embeddings are used at inference. The change in a frozen detector's score between the full-graph and a masked reconstruction is the audit's reading of its dependence on the masked semantics. Node labels are illustrative.}
    \label{fig:etu_overview}
\end{figure*}

\section{Methodology}
ETU turns ``which semantics does this score depend on?'' into an intervention experiment in the feature space of a frozen detector $\Phi$ that assigns anomaly scores to a feature sequence $\mathbf{x} \in \mathbb{R}^{T \times D}$. To do so, for each video $x$ we first generate a hierarchical graph $\mathcal{G}_x$ composed of root nodes (agents in the video), action nodes (the actions the agents take) and outcome nodes (the outcomes of the actions taken by the agents). $\mathcal{G}_x$ is generated by iteratively prompting a VLM teacher (Qwen3-VL~\cite{bai2025qwen3vl}) with the video and three structured prompts (Figure~\ref{fig:teacher_prompts} of the appendix).  
ETU maps $\mathbf{x}$ to a discrete semantic dependency graph $\hat{\mathcal{G}}_x$ (Section~\ref{sec:gem}), a prediction supervised by the teacher graph $\mathcal{G}_x$. ETU then regenerates $\mathbf{x}$ conditioned on $\hat{\mathcal{G}}_x$ or on a masked variant $m(\hat{\mathcal{G}}_x)$ (Section~\ref{sec:cgd}), and reads $\Phi$'s score change between the two reconstructions as the audit's measurement.  

\subsection{Graph Embedding Model (GEM)}
\label{sec:gem}
GEM maps video features $\mathbf{X} \in \mathbb{R}^{T \times D}$ to a structured, intervention-sensitive latent $\mathbf{z}_{\text{graph}} \in \mathbb{R}^{512}$, jointly inferring discrete atoms, their dependency edges and a global structural prior.

\textbf{Node Reconstruction.} We map video features to human-verifiable concepts by allocating $M=25$ node slots for each semantic type, roots ($A_r$), actions ($A_a$), and outcomes ($A_o$), each quantised against a codebook of $N=32$ prototypes. Video features are queried via type-specific cross-attention blocks applied sequentially, yielding continuous hidden representations $\mathbf{n} \in \mathbb{R}^D$. These hidden states are grounded in a discrete manifold via three disjoint codebooks~\cite{van2017neural} $\mathbf{C} \in \mathbb{R}^{N \times D}$. We initialize and regularize each codebook to be orthonormal across the first $D/2$ dimensions, ensuring that each of the $N$ prototypes represents a distinct semantic concept and keeping prototypes composable. A Gumbel-Softmax~\cite{jang2016categorical} estimator computes the categorical selection vector $\mathbf{\hat{n}} \in \mathbb{R}^N$ over the prototypes. The atom representation is $\bar{\mathbf{n}} = \hat{\mathbf{n}}\mathbf{C} + \text{MLP}([\hat{\mathbf{n}}\mathbf{C} \parallel \mathbf{n}])$ for the first 3K training steps and $\bar{\mathbf{n}} = \hat{\mathbf{n}}\mathbf{C} + \text{MLP}(\hat{\mathbf{n}}\mathbf{C})$ afterwards, so that at inference each atom depends only on its selected prototype. To align these latent nodes with the extracted semantic atoms, Hungarian matching~\cite{hungarian} grounds each $\bar{\mathbf{n}}$ to the text embedding $\mathbf{e}$ of a VLM-derived atom via a reconstruction loss:
\begin{equation}
    \mathcal{L}_{\text{recon}} = \sum_{t \in \{r,a,o\}} \sum_{(k,j) \in \pi_t} ( 1 - \cos(\bar{\mathbf{n}}_k, \mathbf{e}_j) + \|\bar{\mathbf{n}}_k - \mathbf{e}_j\|_2^2 ),
\end{equation}
where $\pi_t$ is the set of atoms of type $t$ generated by the VLM and assigned to the $M$ slots of $\hat{\mathcal{G}}_x$ allocated for that type.
Furthermore, to prevent prototype collapse and preserve the partially-orthonormal structure required for composable concepts, we regularize the codebook with $\mathcal{L}_{\text{codebook}}$. This loss minimizes inter-prototype correlation and penalizes the deviation of the Gram matrix from the identity $\mathbf{I} \in \mathbb{R}^{N \times N}$:
\begin{equation}
\begin{aligned}
\mathcal{L}_{\text{codebook}} = {} & \frac{1}{N(N-1)} \sum_{i \neq j} \cos(\mathbf{c}_i, \mathbf{c}_j) \\
& + \frac{1}{N^2} \big\| \mathbf{C}_{:,1:D/2} \mathbf{C}_{:,1:D/2}^\top - \mathbf{I} \big\|_F^2
\end{aligned}
\end{equation}
where $\|\cdot\|_F$ denotes the Frobenius norm.

\textbf{Edge Prediction.} Dependency edges are predicted hierarchically. Each root attends over the action nodes, and the Root-Action adjacency $\mathbf{A}_{RA} \in [0,1]^{M \times M}$ is a sigmoid over an MLP applied to each (root, action) pair; each action node is then updated with its root parents, $\mathbf{a}_{\text{ref}}^{(i)} = \text{MLP}([\mathbf{a}^{(i)} \parallel \sum_{k} \mathbf{A}_{RA}^{(k,i)} \mathbf{r}^{(k)}])$, and a pairwise MLP scores each Action-Outcome edge $\mathbf{A}_{AO}$ from the two nodes, their absolute difference and their Hadamard product (Section~\ref{subsec:edge_predictor} of the appendix).

To handle graph sparsity, we optimize the total edge loss $\mathcal{L}_{\text{edge}} = \mathcal{L}_{RA} + \mathcal{L}_{AO}$ using Online Hard Example Mining (OHEM)~\cite{shrivastava2016training} to retain all positives and the three hardest negatives per positive. For each adjacency matrix $\mathbf{A}_{ST} \in \{\mathbf{A}_{RA}, \mathbf{A}_{AO}\}$, the loss is defined as:
\begin{equation}
\begin{split}
    \mathcal{L}_{ST} = & \frac{1}{N_{\text{sel}}} \sum_{(i,j) \in \mathcal{E}^+} \text{BCE}(\mathbf{A}_{ST}^{(i,j)}, 1) \\
    & + \frac{1}{N_{\text{sel}}}\sum_{(i,j) \in \mathcal{E}^{-}_{\text{hard}}} \text{BCE}(\mathbf{A}_{ST}^{(i,j)}, 0)
\end{split}
\end{equation}
where $N_{\text{sel}}$ counts the mined samples, $\mathcal{E}^+$ the positives and $\mathcal{E}^{-}_{\text{hard}}$ the hard negatives. For the sparser action-outcome transitions, we up-weight the positive term of $\mathcal{L}_{AO}$ by a factor of 5.
A degree-sparsity hinge then prunes redundant correlations:
\begin{equation}
    \mathcal{L}_{\text{sparsity}} = \frac{1}{|\mathcal{V}_{src}|} \sum_{i \in \mathcal{V}_{src}} \max(0, \sum_{j \in \mathcal{V}_{tgt}} \mathbf{A}_{i,j} - \tau)
\end{equation}
where $\mathcal{V}_{tgt} = A_a$ if $\mathcal{V}_{src} = A_r$ and $\mathcal{V}_{tgt} = A_o$ if $\mathcal{V}_{src} = A_a$. With $\tau=1$, the same hinge is also applied to the in-degree of every target node, penalizing nodes with more than one expected edge.

\textbf{Graph Pooling.} The graph is compressed into a global latent $\mathbf{z}_{\text{graph}} \in \mathbb{R}^{512}$ via a gating mechanism that weights each atom by its structural importance. For each node $i$, we compute $w_i = \sigma(\text{MLP}([\mathbf{n}_i \parallel \log(1 + d_i)]))$, where $d_i = \sum_j \mathbf{A}_{i,j}$ is the aggregate out-degree of node $i$. The resulting pooled embedding $w$ is concatenated with a structural prior $\mathbf{s}_{\text{edge}} =  [ \sum_{i,j} (A_{ij} > 0.5), \frac{1}{M^2} \sum_{i,j} A_{ij}, \frac{\sum_{i,j} (A_{ij} > 0.5)}{M(M-1)} ]^\top$ and projected to obtain the graph latent $\mathbf{z}_{\text{graph}}$.

To shape the latent manifold for counterfactual auditing, the model minimizes a contrastive loss. As shown in Figure~\ref{fig:etu_overview}, targeted masking of the reconstructed graph allows us to generate three counterfactual latents, $\mathbf{z}_{\text{irr}}$ (removal of nodes that are not on the path to an outcome node), $\mathbf{z}_{\text{rel}}$ (removal of nodes that are on the path of an outcome node), and $\mathbf{z}_{\text{sem}}$ (substitution of nodes closest to the ground-truth atoms with randomly-sampled, factually incorrect atoms). The teacher lists ordinary and unusual events alike, so the relevant mask removes every outcome chain of a video and not only the anomalous one. The loss enforces structural invariance to noise while maximizing sensitivity to semantic interventions:
\begin{equation}
\label{eq:gem_cont}
\begin{split}
    \mathcal{L}_{\text{cont}}^{\text{GEM}} = & \lambda_\text{irr}(1 - \cos(\mathbf{z}_{\text{graph}}, \mathbf{z}_{\text{irr}})) \\
    & + \lambda_\text{sem}\max(0, \cos(\mathbf{z}_{\text{graph}}, \mathbf{z}_{\text{sem}})) \\
    & + \lambda_\text{rel} \max(0, \cos(\mathbf{z}_{\text{graph}}, \mathbf{z}_{\text{rel}}))
\end{split}
\end{equation}
Finally, VICReg regularization~\cite{bardes2021vicreg} prevents embedding collapse by enforcing unit variance, minimizing off-diagonal covariance ($Cov$), and ensuring invariance under irrelevant masking:
\begin{equation}
\label{eq:gem_vicreg}
\begin{split}
    \mathcal{L}_{\text{vicreg}} = & \frac{1}{D} \sum_{d=1}^D \max(0, 1 - \sqrt{Var(\mathbf{z}_{\text{graph}}^{(d)})}) \\
    & + \frac{1}{D} \sum_{d \neq d'} [Cov(\mathbf{z}_{\text{graph}})]_{d,d'}^2 + \text{MSE}(\mathbf{z}_{\text{graph}}, \mathbf{z}_{\text{irr}}).
\end{split}
\end{equation}
The first term of Eq.~\ref{eq:gem_cont} and the last term of Eq.~\ref{eq:gem_vicreg} pull $\mathbf{z}_{\text{irr}}$ onto $\mathbf{z}_{\text{graph}}$, and on normal videos the last term of Eq.~\ref{eq:gem_cont} pulls $\mathbf{z}_{\text{rel}}$ onto it as well, so this loss uses the video-level label of the training videos. A near-zero $\Delta_{\text{irr}}$ is therefore a result of training and does not describe the detector, and ETU does not measure a dependence on background content. Section~\ref{sec:controls} tests the divergence objective of Eq.~\ref{eq:gem_cont} by removing it from GEM and by adding one to the generator. Implementation details are given in Section~\ref{sec:supp_gem} of the appendix.

\subsection{Graph-Conditioned Feature Diffusion (GFD)}
\label{sec:cgd}
The second stage of ETU employs a Graph-Conditioned Feature Diffusion (GFD) model to denoise continuous video features conditioned on the semantic graph latent $\mathbf{z}_{\text{graph}}$ and its masked variants ($\mathbf{z}_{\text{irr}}, \mathbf{z}_{\text{rel}}, \mathbf{z}_{\text{sem}}$). Because the denoiser is conditioned on the graph, a masked graph changes the reconstruction only through what the conditioning no longer encodes.

\textbf{Architecture.} GFD processes the noisy temporal features $\mathbf{x}_t \in \mathbb{R}^{T \times D_{\text{feat}}}$ using a 4-block Diffusion Transformer (DiT)~\cite{peebles2023scalable}. GFD conditions twice within each transformer block, so that the global topology guides the temporal denoising trajectory. The graph latent $\mathbf{z}_{*} \in \{\mathbf{z}_{\text{graph}}, \mathbf{z}_{\text{irr}}, \mathbf{z}_{\text{rel}}, \mathbf{z}_{\text{sem}}\}$ modulates the features via Feature-wise Linear Modulation (FiLM)~\cite{perez2018film}, and is projected into four latent tokens that serve as the $K$ and $V$ in a cross-attention layer following the temporal self-attention. All video features are normalized by $1^{\text{st}}$--$99^{\text{th}}$ percentile clipping and per-sample standardization ($\mu=0, \sigma=1$), which keeps the denoiser from collapsing to an unconditional prediction.

\begin{table*}[ht]
    \centering
    \small
    \caption{\textbf{Regenerating under the complete graph costs the detectors very little, while substituting the scene semantics or removing the outcome chain costs every one of them.} $\Delta_{\text{graph}}$ is the drop from the clean features to the full-graph reconstruction; $\Delta_{\text{rel}}$, $\Delta_{\text{irr}}$ and $\Delta_{\text{sem}}$ are the reconstruction-referenced deltas of Eq.~\ref{eq:delta}, in frame-AUC points on UCF-Crime and AP points on XD-Violence, positive for a drop; the largest masked delta of each detector on each track is in bold.}
    \label{tab:audit_final}
    \setlength{\tabcolsep}{2.6pt}
    \fitcol{\begin{tabular*}{\textwidth}{@{\extracolsep{\fill}}l cccc cccc cccc cccc}
        \toprule
        \multirow{3}{*}{\textbf{Detector}} & \multicolumn{8}{c}{\textbf{I3D}} & \multicolumn{8}{c}{\textbf{CLIP}} \\
        \cmidrule(lr){2-9} \cmidrule(lr){10-17}
        & \multicolumn{4}{c}{\textbf{UCF-Crime (AUC)}} & \multicolumn{4}{c}{\textbf{XD-Violence (AP)}} & \multicolumn{4}{c}{\textbf{UCF-Crime (AUC)}} & \multicolumn{4}{c}{\textbf{XD-Violence (AP)}} \\
        \cmidrule(lr){2-5} \cmidrule(lr){6-9} \cmidrule(lr){10-13} \cmidrule(lr){14-17}
        & $\Delta_{\text{graph}}$ & $\Delta_{\text{rel}}$ & $\Delta_{\text{irr}}$ & $\Delta_{\text{sem}}$ & $\Delta_{\text{graph}}$ & $\Delta_{\text{rel}}$ & $\Delta_{\text{irr}}$ & $\Delta_{\text{sem}}$ & $\Delta_{\text{graph}}$ & $\Delta_{\text{rel}}$ & $\Delta_{\text{irr}}$ & $\Delta_{\text{sem}}$ & $\Delta_{\text{graph}}$ & $\Delta_{\text{rel}}$ & $\Delta_{\text{irr}}$ & $\Delta_{\text{sem}}$ \\
        \midrule
        MGFN & {$3.25$} & {$6.93$} & {$0.68$} & {$\mathbf{10.02}$} & {$0.97$} & {$39.30$} & {$3.54$} & {$\mathbf{40.31}$} & $0.09$ & $6.78$ & $-0.01$ & {$\mathbf{37.49}$} & {$1.24$} & {$2.90$} & {$0.37$} & {$\mathbf{13.27}$} \\
UR-DMU & {$1.70$} & {$12.35$} & {$-0.65$} & {$\mathbf{14.40}$} & {$0.22$} & {$27.59$} & {$2.21$} & {$\mathbf{40.09}$} & $0.13$ & $\mathbf{5.21}$ & $-0.02$ & {$4.03$} & {$0.71$} & {$2.58$} & {$-0.13$} & {$\mathbf{8.95}$} \\
RTFM & {$1.93$} & {$9.08$} & {$-0.41$} & {$\mathbf{11.98}$} & {$0.70$} & {$29.14$} & {$2.76$} & {$\mathbf{53.62}$} & $-0.10$ & $6.77$ & $0.00$ & {$\mathbf{16.48}$} & {$0.86$} & {$1.32$} & {$0.14$} & {$\mathbf{23.41}$} \\
PEL4VAD & {$1.10$} & {$\mathbf{11.19}$} & {$0.69$} & {${10.00}$} & {$0.05$} & {$16.71$} & {$1.83$} & {$\mathbf{33.09}$} & -- & -- & -- & -- & -- & -- & -- & -- \\
DSANet & -- & -- & -- & -- & -- & -- & -- & -- & $0.16$ & $4.23$ & $-0.01$ & {$\mathbf{37.52}$} & {$0.65$} & {$\mathbf{39.47}$} & {$0.02$} & {${27.86}$} \\
VadCLIP & -- & -- & -- & -- & -- & -- & -- & -- & $0.03$ & $6.47$ & $-0.03$ & {$\mathbf{22.58}$} & {$0.83$} & {$\mathbf{30.42}$} & {$-0.30$} & {${23.63}$} \\
GlanceVAD & {$1.66$} & {$9.21$} & {$0.65$} & {$\mathbf{10.09}$} & {$0.20$} & {$30.65$} & {$1.86$} & {$\mathbf{43.42}$} & $0.30$ & $5.09$ & $-0.01$ & {$\mathbf{31.75}$} & {$0.02$} & {$1.58$} & {$0.02$} & {$\mathbf{9.54}$} \\

        \bottomrule
    \end{tabular*}}
\end{table*}

\textbf{Training.} GFD is trained to reconstruct the encoder features under the complete graph and nothing else. A noise vector $\varepsilon \sim \mathcal{N}(\mathbf{0}, \mathbf{I})$ and a timestep $t \in [0, T]$ are drawn per sample, and the loss optimizes the noise-prediction space and the feature space jointly:
\begin{equation}
\label{eq:cgd_diff}
    \mathcal{L}_{\text{diff}} = \|\varepsilon - \varepsilon_{\text{graph}}\|_2^2 + \|\mathbf{x}_0 - \hat{\mathbf{x}}_{\text{graph}}\|_2^2
\end{equation}
where $\varepsilon_{\text{graph}}$ is the noise predicted under $\mathbf{z}_{\text{graph}}$ and $\hat{\mathbf{x}}_{\text{graph}}$ is the clean feature estimated via Tweedie's formula~\cite{rout2024beyond}. The masked latents $\mathbf{z}_{\text{irr}}$, $\mathbf{z}_{\text{rel}}$ and $\mathbf{z}_{\text{sem}}$ never enter the loss.

\textbf{Inference.} During evaluation, we leverage a partial-reverse DDIM~\cite{song2020denoising} protocol at an intermediate noise level ($t{=}300$ of $T{=}1000$ unless stated). Ground-truth features $\mathbf{x}_0$ are forward-diffused to timestep $t$ to obtain the noisy counterpart $\mathbf{x}_t$. We then perform a 10-step deterministic DDIM reverse process conditioned on each variant $\mathbf{z}_* \in \{\mathbf{z}_{\text{graph}}, \mathbf{z}_{\text{irr}}, \mathbf{z}_{\text{rel}}, \mathbf{z}_{\text{sem}}\}$. To isolate the influence of the graph structure while controlling for reconstruction bias, we compute the counterfactual anomaly delta $\Delta_{m}(\mathbf{x}_0)$ as the change in the detector's benchmark score relative to the full-graph reconstruction:
\begin{equation}
\label{eq:delta}
    \Delta_{m}(\mathbf{x}_0) =  \Phi(\hat{\mathbf{x}}_{\text{graph}}) -  \Phi(\hat{\mathbf{x}}_m)
\end{equation}
where $\Phi(\cdot)$ denotes the frozen VAD model under audit. Every delta in this paper is a difference of dataset-level benchmark points between the two conditionings, frame-AUC on UCF-Crime and AP on XD-Violence, where a positive value means that the mask lowered the performance. It is invariant to a monotone rescaling of the scores, so it reports a change in the ranking of the frames. A high delta indicates sensitivity to the mask $m$, and a near-zero delta invariance. Each mask removes a node together with every node that depends on it downstream in the graph, and the audit describes a detector over a dataset and not on a single video.

\section{Experiments}
\label{sec:experiments}

We audit seven frozen detectors~\cite{AAAI23MGFN, AAAI23URDMU, tian2021weakly, pu2024learning, zhang2024glancevad, yin2026dsanet, AAAI2024_vadclip}, but ETU can be seamlessly employed to audit any VAD-model for which it is possible to reconstruct the latent space of the training data. We carry out this audit on UCF-Crime~\cite{sultani2018real}, which contains real-world CCTV footage, and XD-Violence~\cite{wu2020not}, which contains movie clips, with the two most used encoders in the field, I3D~\cite{i3d} and CLIP~\cite{CLIP}. PEL4VAD is audited on I3D features only, DSANet and VadCLIP on CLIP features only. We follow the standard evaluation protocols of the two datasets~\cite{CVPR23UMIL, AAAI2024_vadclip, sultani2018real, wu2020not} and report every delta as the reconstruction-referenced change of Eq.~\ref{eq:delta}, at noise level $t{=}300$ unless stated otherwise. Every configuration is trained with three seeds, and we report the mean over seeds. Configurations are selected on a held-out subset of the training videos, never on the test videos. No VLM is used at test time. GEM predicts the graph of a test video from its features alone through the learned codebooks, and every mask is computed on that predicted graph. The experiments of Section~\ref{sec:controls} are carried out on UCF-Crime with CLIP features. Two further chains serve as ablations on the same track, namely a \emph{divergence-trained generator}, the same architecture trained with an added divergence objective on its reconstructions (Section~\ref{subsec:divergence_generator} of the appendix), and a \emph{divergence-free ETU}, in which GEM is trained without the objective of Eq.~\ref{eq:gem_cont}.

\subsection{Auditing Frozen Detectors}
\label{sec:audit}
Table~\ref{tab:audit_final} reports the audit. Its first column per track, $\Delta_{\text{graph}}$, shows that regenerating the features under the complete graph costs the detectors very little. On UCF-Crime with CLIP features the drop lies between $\FidelityRegenLo$ and $\FidelityRegenHi$ AUC points, and across the four tracks it is below one point on 16 of the 22 track-detector cells and at most 3.25 points, for MGFN on UCF-Crime with I3D features, so the audited models keep their operating points on regenerated input, and every masked delta is read against a reconstruction that the detectors barely notice. The clean scores are reported in Table~\ref{tab:vad_gt_graph_wrap} of the appendix. On UCF-Crime/CLIP, substituting the scene semantics results in the largest drop for most detectors, removing the outcome chain costs each of the six between \FidelityGateTthree{} and \ReconRelMaxTthree{} AUC points, and pruning the background nodes costs nothing beyond the regeneration. For scale, deleting the 10\% of feature dimensions with the largest magnitude costs the same detectors between 3.50 and 15.37 AUC points (Table~\ref{tab:baselines} of the appendix). Under the relevant mask the six detectors respond by similar amounts, the largest delta exceeding the smallest by a factor of \SpreadFidelityTthree, while the six clean scores lie within \CleanAucSpanUcfClip{} points of one another. On the other three tracks as well, substituting the scene semantics or removing the outcome chain costs far more than pruning the background nodes, which costs nothing beyond the regeneration on CLIP and at most 3.54 AP points on I3D. VLM explainers such as Holmes-VAU~\cite{zhang2025holmes} are not in Table~\ref{tab:audit_final} because their scores do not read the feature space that the audit intervenes on, and counterfactual prompting does not replace the audit, since their score stays nearly unchanged under a counterfactual prompt whether the explanation changes or not (Figure~\ref{fig:holmes1}; Section~\ref{sec:vlm_pitfall} of the appendix).

\subsection{What the Delta Measures}
\label{sec:controls}
In this section, we analyze what the delta of Eq.~\ref{eq:delta} responds to. A change of score under a masked reconstruction admits explanations other than the semantics that were removed, namely the cost of regenerating the features, the size of the perturbation, and the training of the generator itself. We test each of them on UCF-Crime with CLIP features, the perturbation size and the displacement for the relevant mask only, and we then ablate the divergence objective in the generator and in GEM to identify what makes the intervention targeted. 

\begin{table}[!t]
    \centering
    \scriptsize
    \caption{\textbf{ETU on UCF-Crime with CLIP features.} Frame-AUC points lost from the full-graph reconstruction under each mask (Eq.~\ref{eq:delta}) at three noise levels $t$ ($\Delta_{\text{irr}}$ at $t{=}300$); $\Delta_{\text{graph}}$ is the cost of the regeneration itself and $\Delta_{\text{uncond}}$ that of regenerating without any conditioning; the default noise level, $t{=}300$, is in bold.}
    \label{tab:ctrl_reconly}
    \setlength{\tabcolsep}{1.6pt}
    \fitcol{\begin{tabular*}{\linewidth}{@{\extracolsep{\fill}}@{}l rr r rrr rrr@{}}
    \toprule
    & & & & \multicolumn{3}{c}{$\Delta_{\text{rel}}$ at $t$} & \multicolumn{3}{c}{$\Delta_{\text{sem}}$ at $t$} \\
    \cmidrule(lr){5-7} \cmidrule(lr){8-10}
    \textbf{Detector} & $\Delta_{\text{graph}}$ & $\Delta_{\text{uncond}}$ & $\Delta_{\text{irr}}$ & 100 & 300 & 600 & 100 & 300 & 600 \\
    \midrule
    MGFN & $0.09$ & {$0.03$} & $-0.01$ & $1.09$ & $\mathbf{6.78}$ & $13.71$ & 4.26 & {$\mathbf{37.49}$} & 38.73 \\
UR-DMU & $0.13$ & {$-0.10$} & $-0.02$ & $0.61$ & $\mathbf{5.21}$ & $14.49$ & 0.43 & {$\mathbf{4.03}$} & 6.43 \\
RTFM & $-0.10$ & {$-0.15$} & $0.00$ & $0.55$ & $\mathbf{6.77}$ & $19.41$ & 3.72 & {$\mathbf{16.48}$} & 22.64 \\
DSANet & $0.16$ & {$-0.01$} & $-0.01$ & $0.40$ & $\mathbf{4.23}$ & $10.79$ & 4.92 & {$\mathbf{37.52}$} & 37.99 \\
VadCLIP & $0.03$ & {$0.05$} & $-0.03$ & $0.61$ & $\mathbf{6.47}$ & $13.84$ & 3.77 & {$\mathbf{22.58}$} & 25.84 \\
GlanceVAD & $0.30$ & {$-0.04$} & $-0.01$ & $0.36$ & $\mathbf{5.09}$ & $12.07$ & 5.12 & {$\mathbf{31.75}$} & 35.54 \\

    \bottomrule
    \end{tabular*}}
\end{table}
\begin{table*}[!t]
    \centering
    \footnotesize
    \caption{\textbf{The count-matched random deletion under the three regimes.} For each test video, a random draw of as many nodes as the relevant mask removes; the cost of the relevant mask, the cost of the random draw and their difference with its 95\% confidence interval, in frame-AUC points at $t{=}300$ on the UCF-Crime/CLIP test videos. The divergence objective is used in GEM only by ETU, in both GEM and GFD by the divergence-trained generator (against its own full-graph reconstruction), and nowhere by the divergence-free ETU. The differences of ETU, positive on all six detectors with intervals that exclude zero, are in bold.}
    \label{tab:regimes}
    \setlength{\tabcolsep}{4pt}
    \fitcol{\begin{tabular*}{\textwidth}{@{\extracolsep{\fill}}@{}l rrl rrl rrl@{}}
    \toprule
    & \multicolumn{3}{c}{\textbf{ETU}} & \multicolumn{3}{c}{\textbf{Divergence-trained generator}} & \multicolumn{3}{c}{\textbf{Divergence-free ETU}} \\
    \cmidrule(lr){2-4} \cmidrule(lr){5-7} \cmidrule(lr){8-10}
    \textbf{Detector} & $\Delta_{\text{rel}}$ & $\Delta_{\text{rand}}$ & $\Delta_{\text{rel}} - \Delta_{\text{rand}}$ & $\Delta_{\text{rel}}$ & $\Delta_{\text{rand}}$ & $\Delta_{\text{rel}} - \Delta_{\text{rand}}$ & $\Delta_{\text{rel}}$ & $\Delta_{\text{rand}}$ & $\Delta_{\text{rel}} - \Delta_{\text{rand}}$ \\
    \midrule

MGFN & {$6.78$} & {$-1.45$} & {$\mathbf{8.23}$ {\scriptsize$[4.89,\,21.36]$}} & {$8.76$} & {$17.95$} & {$-9.20$ {\scriptsize$[-24.41,\,11.57]$}} & {$\DivFreeRelMgfn$} & {$\DivFreeRandMgfn$} & {$\DivFreeDiffMgfn$ {\scriptsize$[\DivFreeCiLoMgfn,\,\DivFreeCiHiMgfn]$}} \\
UR-DMU & {$5.21$} & {$-0.68$} & {$\mathbf{5.89}$ {\scriptsize$[3.98,\,15.27]$}} & {$-2.10$} & {$9.88$} & {$-11.98$ {\scriptsize$[-28.93,\,-3.91]$}} & {$\DivFreeRelUrdmu$} & {$\DivFreeRandUrdmu$} & {$\DivFreeDiffUrdmu$ {\scriptsize$[\DivFreeCiLoUrdmu,\,\DivFreeCiHiUrdmu]$}} \\
RTFM & {$6.77$} & {$0.86$} & {$\mathbf{5.91}$ {\scriptsize$[2.93,\,18.67]$}} & {$-0.46$} & {$42.48$} & {$-42.93$ {\scriptsize$[-60.55,\,-26.20]$}} & {$\DivFreeRelRtfm$} & {$\DivFreeRandRtfm$} & {$\DivFreeDiffRtfm$ {\scriptsize$[\DivFreeCiLoRtfm,\,\DivFreeCiHiRtfm]$}} \\
DSANet & {$4.23$} & {$-1.22$} & {$\mathbf{5.45}$ {\scriptsize$[3.22,\,14.21]$}} & {$42.81$} & {$13.53$} & {$29.27$ {\scriptsize$[0.89,\,50.50]$}} & {$\DivFreeRelDsanet$} & {$\DivFreeRandDsanet$} & {$\DivFreeDiffDsanet$ {\scriptsize$[\DivFreeCiLoDsanet,\,\DivFreeCiHiDsanet]$}} \\
VadCLIP & {$6.47$} & {$1.13$} & {$\mathbf{5.34}$ {\scriptsize$[0.47,\,22.76]$}} & {$32.75$} & {$12.58$} & {$20.17$ {\scriptsize$[-6.98,\,44.59]$}} & {$\DivFreeRelVadclip$} & {$\DivFreeRandVadclip$} & {$\DivFreeDiffVadclip$ {\scriptsize$[\DivFreeCiLoVadclip,\,\DivFreeCiHiVadclip]$}} \\
GlanceVAD & {$5.09$} & {$0.37$} & {$\mathbf{4.72}$ {\scriptsize$[4.45,\,19.23]$}} & {$9.31$} & {$9.56$} & {$-0.25$ {\scriptsize$[-7.44,\,8.81]$}} & {$\DivFreeRelGlancevad$} & {$\DivFreeRandGlancevad$} & {$\DivFreeDiffGlancevad$ {\scriptsize$[\DivFreeCiLoGlancevad,\,\DivFreeCiHiGlancevad]$}} \\

    \bottomrule
    \end{tabular*}}
\end{table*}
\textbf{Regeneration cost.} Table~\ref{tab:ctrl_reconly} shows that the deltas are not explained by the cost of regeneration. Regenerating the features under the complete graph costs the detectors between $\FidelityRegenLo$ and $\FidelityRegenHi$ AUC points, regenerating them without any conditioning changes the AUC by at most \CtrlUncondFloor{} because it starts from the noised original features, and pruning the background nodes costs nothing beyond the regeneration, whereas removing the outcome chain costs between \FidelityGateTthree{} and \ReconRelMaxTthree{} points on the same weights. We further observe that the relevant-mask and semantic-mask deltas grow monotonically with the noise level at which the regeneration starts, as an effect of the conditioning should. The deltas therefore come from the conditioning and not from the regeneration.

\textbf{Perturbation size.} To test whether any set of nodes of the same size costs the same, we draw for each video a random set of as many nodes as the relevant mask removes and regenerate the features under it. As shown in Table~\ref{tab:regimes} (left), the random deletion has a delta between $\RandCtrlRandLo$ and $\RandCtrlRandHi$ points, whereas removing the outcome chain costs the detectors between \RandCtrlRelLo{} and \RandCtrlRelHi{} points, and the difference is positive for all six detectors with confidence intervals that exclude zero (the resampling procedure is described in Section~\ref{sec:supp_intervals} of the appendix).

\begin{table}[!t]
    \centering
    \scriptsize
    \caption{\textbf{Displacement and detector response under the relevant mask and the count-matched random deletion.} Median cosine and MSE of each masked reconstruction to the full-graph one, next to the frame-AUC drop of Eq.~\ref{eq:delta} averaged over the six detectors. The last row is the median of the per-video ratios random$/$relevant of $1-\cos$ and of MSE, which differs from the ratio of the medians above, and the ratio of the mean deltas. Bold marks the larger displacement and response over all videos.}
    \label{tab:displacement}
    \setlength{\tabcolsep}{3pt}
    \fitcol{\begin{tabular*}{\linewidth}{@{\extracolsep{\fill}}@{}l rr r@{}}
    \toprule
    & \multicolumn{2}{c}{\textbf{Displacement}} & \multicolumn{1}{c}{\textbf{Detector response}} \\
    \cmidrule(lr){2-3} \cmidrule(lr){4-4}
    \textbf{Mask} & $\cos$ & MSE & $\Delta$ (AUC points) \\
    \midrule
    irrelevant mask & {$1.000$} & {$0.000$} & {$-0.01$} \\
relevant mask & {$0.977$} & {$0.009$} & {$\mathbf{5.76}$} \\
\quad anomalous videos & {$0.983$} & {$0.006$} & {$10.46$} \\
count-matched random & {$\mathbf{0.955}$} & {$\mathbf{0.020}$} & {$-0.17$} \\
\quad anomalous videos & {$0.989$} & {$0.005$} & {$-0.43$} \\
\midrule
random$/$relevant ratio & {$0.95$} & {$1.11$} & {$-0.03$} \\

    \bottomrule
    \end{tabular*}}
\end{table}
\textbf{Displacement.} Table~\ref{tab:displacement} reports how far each of the two deletions moves the reconstruction from the full-graph one, next to the response of the detectors, on the videos of Table~\ref{tab:regimes}. The median random draw moves the reconstruction farther than the median relevant mask and about as far on anomalous videos specifically, yet the relevant mask costs the detectors 5.76 AUC points on average and the random draw $-0.17$. On the normal videos the random draw displaces the reconstruction farther, while the relevant mask shifts the raw detector scores \SplitNormalShiftRatioLo{} to \SplitNormalShiftRatioHi{} times more (Tables~\ref{tab:displacement_supp} and~\ref{tab:intervals} of the appendix). On the anomalous videos the relevant mask displaces the reconstruction slightly more and costs every detector far more. The conditioning latent shows the same pattern in cosine. On the anomalous videos the random latent is as far from $\mathbf{z}_{\text{graph}}$ as the relevant one ($-0.307$ against $-0.331$), while in $L_2$ the relevant latent is farther (Table~\ref{tab:latent_displacement} of the appendix). We conclude that the detectors respond to which nodes the mask removes, and not to how far it moves the features.

\textbf{Divergence objective in GFD.} To see what changes when the generator is trained to produce the contrast that the audit measures, we train the same architecture with the divergence objective of Eq.~\ref{eq:cgd_cont} of the appendix on the same GEM latent, which keeps the irrelevant-mask reconstruction on the full-graph one and pushes the relevant- and semantic-mask reconstructions at least a margin away from it. As shown in Table~\ref{tab:regimes}, this divergence-trained generator fails the test that ETU passes. The difference between the relevant mask and the random deletion is positive for \QthirtyfourDpos{} of the six detectors, indistinguishable from zero for \QthirtyfourDzero{} and negative for \QthirtyfourDneg. Table~\ref{tab:divgen_states} of the appendix shows why. One random node that it has not seen in training costs it between \CtrlRandUntrainedLo{} and \CtrlRandUntrainedHi{} points, on average 9.91 points against 15.18 for the whole relevant mask, and once a random-node variant is added to its training one random node changes the AUC by at most \CtrlRandTrained{} points (Table~\ref{tab:ctrl_randnode}). Its relevant-mask reconstruction of the anomalous videos is nearly orthogonal to its full-graph reconstruction (median cosine 0.051), while the random draw moves farther still ($-0.217$; Table~\ref{tab:displacement_supp}), and for four of the six detectors the random draw also costs more. A generator trained to separate the interventions thus reacts to any conditioning it was not trained on, and its deltas cannot be attributed to the content that the mask removes; without such an objective, a change in the scores of the audited detectors can only come from the change in the conditioning that GFD receives.

\textbf{Divergence objective in GEM.} ETU still relies on a graph embedding trained with the divergence objective of Eq.~\ref{eq:gem_cont}. To verify that this objective is what makes the intervention effective, we retrain GEM without it and train a new reconstruction-only generator on its latent. As shown in Table~\ref{tab:regimes}, removing the outcome chain and removing the same number of random nodes then have the same effect on all six detectors, and both effects are small, although the new GEM reconstructs the graphs as accurately as the original one.

Table~\ref{tab:regimes} therefore shows three regimes of the same intervention. With the divergence objective in GEM only, removing the outcome chain moves every detector while a random deletion of the same size moves none, although it displaces the features about as much. With the objective in the generator as well, a random deletion pushes the reconstructed features off the manifold, and the detectors react to the displacement instead of the content that was removed. Without the objective, neither deletion moves the detectors. We conclude that the divergence objective in GEM is the design choice that turns a masked graph into a targeted intervention on the input features, whose effect depends on which nodes are masked and not on the size of the perturbation.

\begin{figure}[!t]
    \centering
    \includegraphics[width=\linewidth]{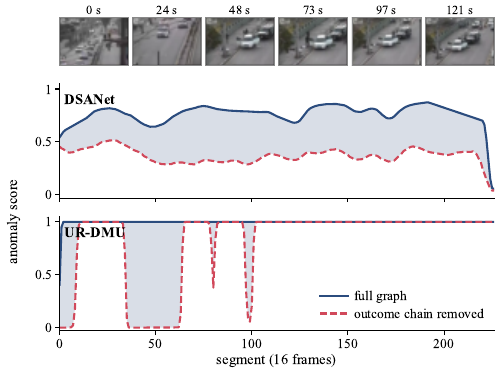}
    \caption{\textbf{The same intervention moves two detectors very differently} (Arrest024, UCF-Crime). Raw anomaly scores of DSANet and UR-DMU under the full graph and with the outcome chain removed (the roots \emph{White pickup truck} and \emph{Police officers}, their collision and arrest actions, and the resulting stop and restraint outcomes). The score of DSANet drops on every segment, whereas UR-DMU is saturated when unmasked and responds all-or-nothing.}
    \label{fig:relmask}
\end{figure}
\textbf{Detector-specific readings.} The audit can also separate detectors whose benchmark scores are close. DSANet and UR-DMU reach clean AUCs of 89.44 and 85.05 on UCF-Crime with CLIP features (Table~\ref{tab:vad_gt_graph_wrap} of the appendix), yet substituting the scene semantics costs DSANet more than UR-DMU (Table~\ref{tab:audit_final}). Figure~\ref{fig:relmask} shows how differently the two detectors respond to the relevant mask on one video. The score of DSANet drops on every segment, whereas UR-DMU is saturated when unmasked and responds all-or-nothing. At the dataset level, masking a single node changes the frame-AUC by at most \FidelityKoneCap{} points when the node lies on the outcome chain and \FidelityKoneOffCap{} points when it lies off it, against \FidelityGateTthree{} to \ReconRelMaxTthree{} points for the whole chain (Table~\ref{tab:single_node} of the appendix). A single-node mask also removes every node that depends on it downstream.

\subsection{Dependence on the Teacher}
\label{sec:teacher}
To assess how much the graphs depend on the VLM that produces them, we parsed the UCF-Crime videos with three teachers, Qwen3-VL (the teacher of all other experiments), InternVideo3 and Gemma-4, using the same prompts, and compared their graphs video by video (Section~\ref{sec:supp_teacher} and Table~\ref{tab:teacher_agreement} of the appendix). The atoms of two teachers on the same video agree more than those of two videos of the same anomaly class on \TeacherPairedWinLo--\TeacherPairedWinHi\% of the videos, whereas the edges agree little, with an edge $F_1$ of \TeacherEdgeFoneLo--\TeacherEdgeFoneHi{} for two teachers on the same video against \TeacherEdgeNullLo--\TeacherEdgeNullHi{} for two videos of the same class. The outcome chain is therefore a property of the teacher's parse, as GEM reproduces it, and not a structure of the scene that any teacher would recover. The parsed graphs do not fill the slots. The mean graph has \MeanRootsPerVideo{} roots, \MeanIntermediatesPerVideo{} actions and \MeanLeavesPerVideo{} outcomes against $M{=}25$ slots per type, while all $N{=}32$ prototypes of each type are in use (Table~\ref{tab:codebook_size} of the appendix).

\section{Conclusion}
\label{sec:conclusion}
We introduced \textit{Explain The Unexplainable} (ETU), a framework that audits frozen Video Anomaly Detection models by regenerating their input features under a masked semantic dependency graph and reading the change in their score, with one set of interventions handed unchanged to every detector. On UCF-Crime/CLIP, substituting the scene semantics results in the largest drop for most detectors, and removing the outcome chain costs every detector more than a count-matched random deletion that moves the reconstruction about as far; the intervention becomes indistinguishable from a random one without the divergence objective of the graph embedding, and a response to the displacement of the reconstruction with a divergence objective in the generator. ETU therefore measures how much a detector's score depends on the graph-conditioned content of its input features. Its generator must stay out of the objective that makes the intervention targeted.

{
    \small
    \bibliographystyle{ieeenat_fullname}
    \bibliography{main}
}

\clearpage
\appendix
\setcounter{table}{0}
\setcounter{figure}{0}
\setcounter{equation}{0}
\renewcommand{\thetable}{S\arabic{table}}
\renewcommand{\thefigure}{S\arabic{figure}}
\renewcommand{\theequation}{S\arabic{equation}}
\renewcommand{\theHtable}{S\arabic{table}}
\renewcommand{\theHfigure}{S\arabic{figure}}
\renewcommand{\theHequation}{S\arabic{equation}}
\section{Method Details and Ablations}

\subsection{Preliminaries on Counterfactual VAD}
\label{sec:preliminaries}
ETU expands the scope of Explainable VAD by shifting the objective from description to intervention. By learning a discrete semantic prior and utilizing guided diffusion to systematically ablate semantic concepts in the latent space, ETU allows to evaluate a pre-trained VAD model via counterfactual interventions. 
We define the VAD task as assigning anomaly scores $S \in \mathbb{R}^{T \times 1}$ to a feature sequence $\mathbf{x} \in \mathbb{R}^{T \times D}$, obtained through a pre-trained video encoder, using a VAD model $\Phi$. To explain the decision logic of $\Phi(\mathbf{x})$, we seek a counterfactual feature reconstruction $\mathbf{\hat{x}}_m$ that represents a minimal, semantically-guided intervention $m$ on the original scene.

\textbf{Semantic Atom Space.} We represent the semantic content of a video as a Directed Acyclic Graph (DAG)~\cite{zheng2018dags}, denoted $\mathcal{G} = (\mathcal{V}, \mathcal{E})$. Following the principles of compositional scene understanding, we define $\mathcal{V}$ as a disjoint set of three semantic atoms, namely \textbf{Root Nodes ($A_r$)} representing agents or structural subjects present in a scene; \textbf{Action Nodes ($A_a$)} representing specific agent behaviors, where directed edges exist such that $(A_r \to A_a) \in \mathcal{E}$; \textbf{Outcome Nodes ($A_o$)} representing resultant states or anomalies, where $(A_a \to A_o) \in \mathcal{E}$. The set of all directed paths $\pi = \{A_r \to A_a \to A_o\}$ forms the outcome chains of a scene. As shown in Figure~\ref{fig:etu_overview}, the ground-truth graph $\mathcal{G}$ is generated by an offline VLM (Qwen3-VL~\cite{bai2025qwen3vl}) observing the video at 1\,fps and emitting atoms and dependency edges directly, following the atomic decomposition of~\cite{teng2025atom}. The topological hierarchy is then strictly enforced to ensure a consistent semantic manifold across the dataset.

\begin{figure}[!t]
    \centering
    \includegraphics[width=0.8\linewidth]{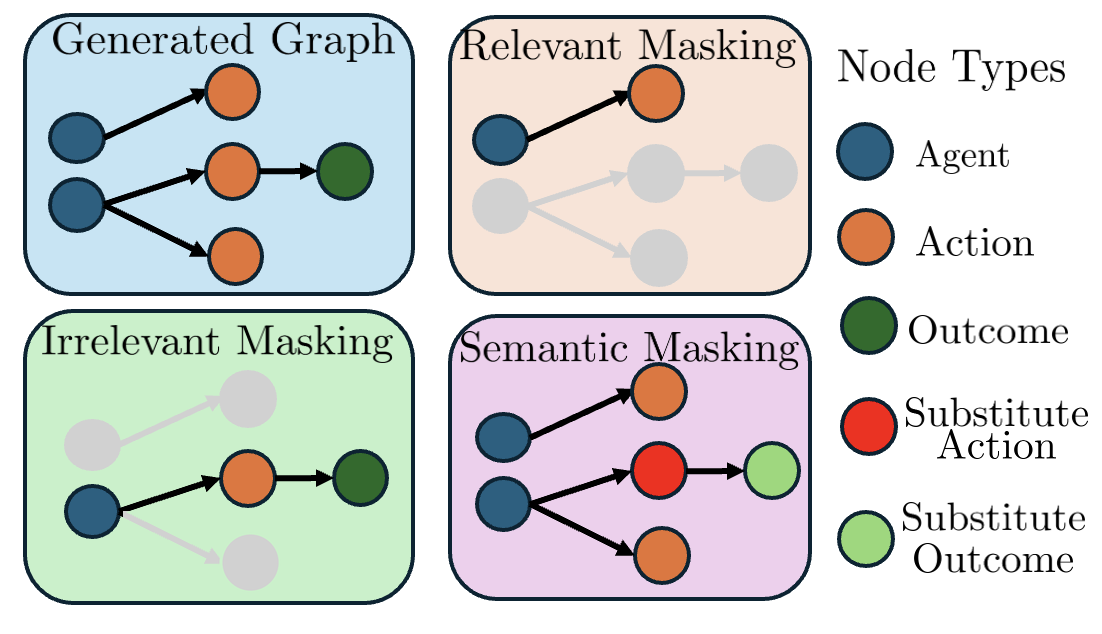}
    \caption{\textbf{Relevant Masking} removes outcome atoms and their upstream dependencies. \textbf{Irrelevant Masking} prunes background elements off every outcome path, preserving the primary anomaly chains. \textbf{Semantic Masking} replaces nodes with concepts from different videos to simulate complete scene substitution.}
    \label{fig:atom_masking}
\end{figure}
\textbf{Counterfactual Interventions.} An intervention $m$ on the semantic manifold, denoted as $m(\mathcal{G})$, involves masking a specific subset of nodes $\mathcal{V}_m \subset \mathcal{V}$. The objective of ETU is to learn a function $h: \mathbf{x} \to \mathcal{G}$ that maps continuous video features to a discrete graph $\mathcal{G}$ that represents the semantic dependency graph of the scene, and then a mapping $k: (\bar{\mathbf{x}}, \mathcal{G}) \to \mathbf{\hat{x}_{m}}$, where $\bar{\mathbf{x}} = \mathbf{x} + \varepsilon \sim \mathcal{N}(\mathbf{0}, \mathbf{I})$, to translate these discrete graph perturbations back into the continuous feature space. The interventions leveraged by ETU are shown in Figure~\ref{fig:atom_masking}. Specifically, we generate a counterfactual feature $\mathbf{\hat{x}}_m = k(\bar{\mathbf{x}}, m(\mathcal{G})) = k(\bar{\mathbf{x}}, m(h(\mathbf{x})))$ by denoising the video features under the masked graph condition. To quantify the sensitivity of the model's decision to the masked atoms, we compute the anomaly delta ($\Delta$) resulting from intervention $m$ against the reconstruction under the unmasked graph, $\mathbf{\hat{x}}_{\text{graph}} = k(\bar{\mathbf{x}}, h(\mathbf{x}))$:
\begin{equation*} \Delta_m = \Phi(\mathbf{\hat{x}}_{\text{graph}}) - \Phi(\mathbf{\hat{x}}_m) \end{equation*}
where a high positive $\Delta$ indicates that the anomaly prediction of model $\Phi$ depended on the masked semantic chain. This formulation allows for three distinct auditing strategies, which remove relevant anomaly chains ($\mathbf{z}_{\text{rel}}$), prune irrelevant background nodes ($\mathbf{z}_{\text{irr}}$), or replace scene semantics ($\mathbf{z}_{\text{sem}}$) to test the robustness of the detector.

\subsection{GEM}
\label{sec:supp_gem}
GEM is trained via a multi-stage curriculum:

\textbf{Implementation details.} GEM receives sequences of 200 CLIP (512-d) or I3D (1024-d) feature snippets and projects them to 512 dimensions. For each node type, $M{=}25$ learned query slots are refined by three cross-attention layers (8 heads) and projected to the 1024-dimensional space of the atom embeddings, in which each teacher label is embedded with Qwen3-Embedding-0.6B. Each type has a codebook of $N{=}32$ prototypes, and the Gumbel-Softmax temperature decreases linearly from 1.0 to 0.5 between steps 500 and 15K. The total loss is $\mathcal{L}_{\text{recon}} + \mathcal{L}_{\text{codebook}} + \mathcal{L}_{\text{edge}} + 0.1\,\mathcal{L}_{\text{sparsity}} + \mathcal{L}^{\text{GEM}}_{\text{cont}} + \mathcal{L}_{\text{vicreg}}$. In $\mathcal{L}_{\text{codebook}}$ the two terms of Eq.~2 of the main paper are weighted by 0.2 each, plus a slot-diversity term (0.5) and an entropy term on the average code usage (2.0); in $\mathcal{L}_{AO}$ the positive term is weighted by 5; in $\mathcal{L}^{\text{GEM}}_{\text{cont}}$, $\lambda_{\text{irr}} = \lambda_{\text{sem}} = \lambda_{\text{rel}} = 0.2$. GEM is trained end to end for 60K steps with AdamW (learning rate $3{\times}10^{-4}$, $0.5{\times}$ for the codebooks, weight decay $10^{-5}$, gradient clipping at 2.0, 1.5K warm-up steps and cosine decay to $10^{-6}$) and batch size 16. The losses are switched on cumulatively, with $\mathcal{L}_{\text{recon}}$ and $\mathcal{L}_{\text{codebook}}$ from step 0, $\mathcal{L}_{\text{edge}}$ from 3K, $\mathcal{L}_{\text{sparsity}}$ and the variance and covariance terms of $\mathcal{L}_{\text{vicreg}}$ from 23K, and $\mathcal{L}^{\text{GEM}}_{\text{cont}}$ and the invariance term of $\mathcal{L}_{\text{vicreg}}$ from 45K.

To isolate the contributions of our design choices, we conducted evaluations on the UCF-Crime dataset. The following subsections detail our findings across node reconstruction, structural edge prediction, and latent space regularization. The selected optimal configuration is highlighted in \colorbox{gray!25}{grey} in the ablation tables.

\subsubsection{Node Reconstruction}
\label{subsec:node_recon}

Given I3D video features $\mathbf{x} \in \mathbb{R}^{T \times 1024}$, GEM learns to reconstruct the three hierarchical levels of a semantic dependency graph, namely root nodes (actors/objects), Action nodes (actions/states), and Outcome nodes (consequences/anomalies). We utilize a discrete codebook of learned prototypes via Gumbel-Softmax selection. We ablate the codebook capacity and regularization.

\subsubsection{Codebook Capacity and Regularization}

Table~\ref{tab:codebook_size} varies the number of prototypes $N$ per node type. All prototypes are used at $N{=}24$ and $N{=}32$; at $N{=}128$ only 20\% to 24\% of each codebook is used. 

\begin{table}[ht]
    \centering
    \caption{Codebook capacity. MSE wt. is the weight of the squared-error term of $\mathcal{L}_{\text{recon}}$ in that run (Table~\ref{tab:mse_weight}); rows 16 and 24 differ from the others in this weight as well as in the number of prototypes.}
    \fitcol{\begin{tabular}{@{}cccccc}
    \toprule
    \textbf{Prototypes} & \textbf{MSE wt.} & \textbf{Test Cosine} & \textbf{Root Util.} & \textbf{Action Util.} & \textbf{Outcome Util.} \\
    \midrule
    16 & 0.2 & 0.591 & 88\% & 100\% & 94\% \\
    24 & 1.0 & 0.734 & 100\% & 100\% & 100\% \\
    \rowcolor{gray!25} \textbf{32} & \textbf{0.5} & \textbf{0.712} & \textbf{100\%} & \textbf{100\%} & \textbf{100\%} \\
    48 & 0.5 & 0.435 & 50\% & 100\% & 42\% \\
    64 & 0.5 & 0.581 & 39\% & 75\% & 69\% \\
    128 & 0.5 & 0.521 & 24\% & 20\% & 23\% \\
    \bottomrule
    \end{tabular}}
    \label{tab:codebook_size}
\end{table}

To further enforce prototype usage, we ablated the Mean Squared Error (MSE) loss weight (Table~\ref{tab:mse_weight}). We found that an MSE weight of 0.5 or greater is needed for full codebook utilization (Table~\ref{tab:mse_weight}); the 16-prototype run of Table~\ref{tab:codebook_size} used 0.2. While higher MSE resulted in better node reconstruction, in the final configuration we set the weight to 0.5 to balance node reconstruction and edge prediction.

\begin{table}[ht]
    \centering
    \caption{Impact of MSE loss weighting on prototype utilization and feature fidelity.}
    \fitcol{\begin{tabular}{@{}cll}
    \toprule
    \textbf{MSE Weight} & \textbf{Utilization} & \textbf{Cosine Sim.} \\
    \midrule
    0.2 & Partial (88\% to 100\%) & 0.34 to 0.59 \\
    \rowcolor{gray!25} \textbf{0.5} & Full (24 to 32 prototypes) & 0.71 to 0.74 \\
    1.0 & Full (24 to 32 prototypes) & 0.73 to 0.81 \\
    \bottomrule
    \end{tabular}}
    \label{tab:mse_weight}
\end{table}

\subsubsection{Edge Prediction}
\label{subsec:edge_prediction}

Given the reconstructed nodes, GEM must predict dependency edges to form a meaningful topology. Due to the extreme class imbalance (valid edges account for approximately \EdgeSparsityPct\% of the \EdgeSlotPairs{} candidate pairs in a reconstructed graph that contains 25 root, 25 action and 25 outcome node slots), robust architectural and optimization choices are required.

\subsubsection{Predictor Architecture and Loss Asymmetry}
\label{subsec:edge_predictor}

We first evaluated pairwise scoring mechanisms (Table~\ref{tab:edge_arch}). A dot product and an MLP on the concatenated pair reached an edge $F_{0.5}$ of 0.18 and 0.22; adding the absolute difference and the element-wise product of the two nodes to the MLP input raised it to 0.29, and hard negative mining (3:1 ratio) and hierarchical conditioning (Root$\to$Action edges informing Action$\to$Outcome edges) raised it to 0.66. In the final model each root attends over the action nodes (4-head cross-attention) and $\mathbf{A}_{RA}^{(k,i)}$ is a sigmoid over an MLP applied to the concatenation of root $k$ and action $i$; each action is refined with its root parents, attends over the outcome nodes, and $\mathbf{A}_{AO}^{(i,j)}$ is a sigmoid over an MLP applied to $[\mathbf{a}^{(i)}_{\text{ref}} \parallel \mathbf{o}^{(j)} \parallel |\mathbf{a}^{(i)}_{\text{ref}}-\mathbf{o}^{(j)}| \parallel \mathbf{a}^{(i)}_{\text{ref}} \odot \mathbf{o}^{(j)}]$.

\begin{table}[ht]
    \centering
    \caption{Ablation of edge scoring architectures and optimization strategies.}
    \fitcol{\begin{tabular}{@{}lcl}
    \toprule
    \textbf{Architecture / Strategy} & \textbf{Edge $F_{0.5}$} & \textbf{Observation} \\
    \midrule
    Dot product & 0.18 & Insufficient capacity for asymmetric edges \\
    MLP pair & 0.22 & Marginal improvement \\
    Cross-attention & 0.25 & Treats pairs independently \\
     MLP with $|\mathbf{a}-\mathbf{o}|$, $\mathbf{a}\odot\mathbf{o}$ & 0.29 & Best baseline scoring mechanism \\
    + Hard negative mining & 0.45 & Addresses class imbalance \\
    \rowcolor{gray!25} + Hierarchical conditioning & \textbf{0.66} & Explicitly models edge directionality \\
    \bottomrule
    \end{tabular}}
    \label{tab:edge_arch}
\end{table}

Furthermore, we observed a bottleneck in predicting Action$\to$Outcome (AO) edges, which lagged behind Root$\to$Action (RA) edges by 5 to 10 $F_{0.5}$ points. This stems from the higher variance in Outcome node embeddings (consequences and anomalies). We corrected this by applying an asymmetric loss weight ($\lambda = 5.0$ for $\mathcal{L}_{AO}$) to false negatives on the AO level, raising the AO $F_{0.5}$ from 0.653 to 0.768.

\subsubsection{Joint Training Dynamics} A challenge in end-to-end training is the destructive interference between node reconstruction and edge prediction. Ground-truth nodes exhibit a 3\% to 6\% cosine separation between connected pairs. However, after passing through the reconstruction bottleneck, this separation drops to 0.02\% (a 275$\times$ reduction), causing modular GEM training to fail entirely ($F_{0.5} \approx 0.01$). 

We solved this via joint end-to-end training, which forces the reconstruction objective to preserve edge-relevant features. To prevent gradient conflicts in shared layers, we implemented dual refinement pathways (separating the edge refinement gradient from the reconstruction gradient), successfully restoring edge prediction capabilities on reconstructed nodes. Detaching the gradients completely (stop-grad) resulted in edge failure ($F_{0.5}=0.323$), confirming that gradients must flow through the shared backbone.

\subsubsection{Graph Embeddings}
\label{subsec:graph_embeddings}

The goal of GEM is to yield a single embedding $z_\text{graph} \in \mathbb{R}^{512}$ that accurately dictates the behavior of the downstream diffusion model. This embedding must be both highly discriminative and counterfactually sensitive.

\subsubsection{Graph Topology Injection}

To aggregate the predicted graph structure into a single vector, we evaluated various pooling and message-passing schemes (Table~\ref{tab:edge_topology}). While standard Graph Isomorphism Networks (GIN)~\cite{xu2018powerful} are popular, the message passing corrupted our specific edge-relevant features, degrading edge $F_{0.5}$ to 0.322. \textbf{Edge-Gated Pooling}, where node representations are weighted by their predicted edge connectivity before mean pooling, safely injected topology into the latent space while maintaining high linear probe performance.

\begin{table}[ht]
    \centering
    \caption{Methods for injecting edge topology into graph embeddings.}
    \fitcol{\begin{tabular}{@{}lccc}
    \toprule
    \textbf{Method} & \textbf{Edge $F_{0.5}$} & \textbf{Linear Probe AUC} & \textbf{CF Gap} \\
    \midrule
    Mean Pooling & 0.690 & 0.893 & 0.969 \\
    GIN (3-layer) & 0.322 & 0.924 & $-$0.004 \\
    Virtual Node & 0.330 & 0.848 & 0.053 \\
    \rowcolor{gray!25} \textbf{Edge-Gated Pooling} & \textbf{0.712} & \textbf{0.861} & \textbf{0.637} \\
    \bottomrule
    \end{tabular}}
    \label{tab:edge_topology}
\end{table}

\subsubsection{Latent Regularization and Counterfactual Training}

To prevent embedding collapse and ensure the latent space reacts correctly to counterfactual manipulations, we use Variance-Invariance-Covariance Regularization (VICReg)~\cite{bardes2021vicreg}. 

As shown in Table~\ref{tab:vicreg_sweep}, applying VICReg with a weight of $\lambda=1.0$ monotonically improves the Linear Probe AUC (LP AUC, measuring linear separability of normal vs. anomalous graphs) and increases the Counterfactual (CF) Gap. The CF gap measures the difference in cosine similarity when the model masks out relevant anomaly nodes versus irrelevant background nodes. A larger gap implies the downstream diffusion model will react more strongly to the removal of anomaly-relevant nodes.

\begin{table}[ht]
    \centering
    \caption{Controlled ablation of VICReg regularization strength ($\lambda$).}
    \fitcol{\begin{tabular*}{\linewidth}{@{\extracolsep{\fill}}cccc}
    \toprule
    \textbf{VICReg $\lambda$} & \textbf{LP AUC} & \textbf{CF Gap} & \textbf{Edge $F_{0.5}$}  \\
    \midrule
    0.00 & 0.843 & 1.031 & 0.697 \\
    0.25 & 0.883 & 0.936 & 0.697 \\
    0.50 & 0.900 & 0.957 & 0.687 \\
    \rowcolor{gray!25} \textbf{1.00} & \textbf{0.932} & \textbf{1.067} & \textbf{0.697} \\
    \bottomrule
    \end{tabular*}}
    \label{tab:vicreg_sweep}
\end{table}

\subsubsection{Curriculum Learning and Phase Transitions}

We observed that the model requires a strict curriculum to converge. The most important finding was the necessity of \textbf{full-phase contrastive training} (Table~\ref{tab:full_phase}). Extending training beyond the joint edge-reconstruction phase to explicitly train the contrastive masking objectives (steps 45K to 60K) yielded improvements in the CF gap (67\%) and general embedding separation, with only a negligible (2\% to 3\%) cost to the structural edge $F_{0.5}$ score.

\begin{table}[ht]
    \centering
    \caption{Impact of extending the curriculum to include full-phase contrastive training.}
    \fitcol{\begin{tabular}{@{}lccc}
    \toprule
    \textbf{Metric} & \textbf{Joint-Phase Only (45K)} & \textbf{Full-Phase (60K)} & \textbf{Relative Change} \\
    \midrule
    CF Gap & 0.60 to 0.64 & 0.94 to 1.07 & \textbf{+55\% to +70\%} \\
    Embed Gap & 0.02 to 0.06 & 0.18 to 0.41 & \textbf{+300\% to +700\%} \\
    LP AUC & 0.86 to 0.88 & 0.88 to 0.93 & +2\% to +5\% \\
    Edge $F_{0.5}$ & 0.71 to 0.73 & 0.69 to 0.70 & $-$2\% to $-$3\% \\
    \bottomrule
    \end{tabular}}
    \label{tab:full_phase}
\end{table}

\subsubsection{Final Selected Configuration and Analysis}
\label{subsec:selected_config}

Based on the explorations detailed above, the selected architecture leverages 32 discrete prototypes, Edge-Gated pooling, VICReg, and asymmetric AO loss weighting, trained through the full 60K step contrastive curriculum. 

\begin{table}[ht]
    \centering
    \caption{Counterfactual masking sensitivity in the final latent space (anomaly videos only).}
    \fitcol{\begin{tabular*}{\linewidth}{@{\extracolsep{\fill}}lc}
    \toprule
    \textbf{Masking Strategy} & \textbf{Cosine Sim.} \\
    \midrule
    Relevant (remove anomaly) & $-$0.123 \\
    Irrelevant (remove background) & 0.944  \\
    Semantic (replace scene) & $-$0.193 \\
    \midrule
    \rowcolor{gray!25} \textbf{CF Gap (Irrel $-$ Rel)} & \textbf{1.067}  \\
    \bottomrule
    \end{tabular*}}
    \label{tab:r20c2_cf}
\end{table}

As shown in Table~\ref{tab:r20c2_cf}, the chosen configuration reaches a Counterfactual Gap of 1.067. When the predicted outcome-chain nodes are masked out (Relevant masking) the embedding cosine similarity drops to $-$0.123, while removing off-chain background nodes leaves the embedding nearly untouched (0.944). This is the behaviour the downstream Graph-Conditioned Feature Diffusion (GFD) model requires of its conditioning; it is a property of the trained embedding, selected for by this very criterion, and not evidence about any audited detector.

\textbf{Divergence-free GEM.} The GEM of the divergence-free ETU (Section~\ref{sec:controls} of the main paper) has the same architecture, data, schedule and seed, with Eq.~\ref{eq:gem_cont} of the main paper removed entirely. Its node reconstruction cosine is \DivFreeEncCos{}, against 0.885 for the GEM of ETU.

\subsubsection{Evaluation and Discussion}
\label{sec:gem_evaluation}
\textbf{GEM.} Table~\ref{tab:comprehensive_pipeline} validates GEM across node reconstruction, edge prediction, and embedding separability. Node reconstruction and edge precision are high on both datasets and backbones, and the embedding separates the masked variants as the downstream GFD requires. The variants $\mathbf{z}_{\text{rel}}$ and $\mathbf{z}_{\text{sem}}$ are near-orthogonal to the full-graph embedding while $\mathbf{z}_{\text{irr}}$ stays close to it. These are properties of the conditioning layer, not of any audited detector. %
\begin{table}[!tb]
    \centering
    \small
    \caption{GEM evaluation across node reconstruction, edge fidelity and embedding similarity.}
    \label{tab:comprehensive_pipeline}
    \setlength{\tabcolsep}{2pt}
    \fitcol{\begin{tabular*}{\linewidth}{@{\extracolsep{\fill}}llcccccc}
        \toprule
        \multirow{2}{*}{\textbf{Dataset}} & \multirow{2}{*}{\textbf{Enc.}} & \multicolumn{2}{c}{\textbf{Node}} & \textbf{Edge} & \multicolumn{3}{c}{\textbf{Graph (Cos.)}} \\
        \cmidrule(lr){3-4} \cmidrule(lr){5-5} \cmidrule(lr){6-8}
        & & Cos. & MSE & $F_{0.5}$ & $\mathbf{z}_{\text{sem}}$ & $\mathbf{z}_{\text{rel}}$ & $\mathbf{z}_{\text{irr}}$ \\
        \midrule
        \multirow{2}{*}{UCF} & I3D & 0.891 & 0.016 & 0.68 & $-0.31$ & 0.25 & 0.96 \\
        & CLIP & 0.885 & 0.017 & 0.64 & $-0.28$ & 0.26 & 0.97 \\
        \midrule
        \multirow{2}{*}{XDV} & I3D & 0.864 & 0.021 & 0.70 & $-0.34$ & 0.15 & 0.99 \\
        & CLIP & 0.862 & 0.020 & 0.68 & $-0.37$ & 0.15 & 0.98 \\
        \bottomrule
    \end{tabular*}}
\end{table}

\textbf{Topological Complexity and Alignment.} High node reconstruction faithfulness ($>0.85$) is consistently achieved, though alignment varies by dataset. UCF-Crime exhibits higher alignment due to the spatio-temporal stability of fixed-camera CCTV footage. Conversely, the movie-based clips in XD-Violence contain frequent temporal discontinuities and high visual variance, resulting in complex ground-truth graphs with higher node density in anomalous segments.

\textbf{Edge Prediction and Metric Selection.} We prioritize precision via the $F_{0.5}$ metric to heavily penalize false-positive edges, as spurious connections would compromise the integrity of the counterfactual audit. GEM achieves $F_{0.5} \ge 0.64$ across all settings. Given the significant class imbalance between candidate and actual edges, this performance demonstrates a robust ability to filter environmental noise while preserving the essential outcome-chain backbone.

\textbf{Embedding Separation.} Graph embedding similarities show that the conditioning layer separates the relevant and semantic variants from the full graph. Irrelevant maskings ($\mathbf{z}_{\text{irr}}$) remain near-identical to the full graph ($\mathbf{z}_{\text{graph}}$), while the relevant ($\mathbf{z}_{\text{rel}}$) and semantic ($\mathbf{z}_{\text{sem}}$) variants diverge, the semantic one to negative cosine similarity, most markedly on XD-Violence. This is the separation that the contrastive objective of Section~\ref{sec:gem} of the main paper trains for. Whether the score of a frozen detector depends on what was removed is established by the experiments of Sections~\ref{sec:audit} and~\ref{sec:controls} of the main paper.

\subsection{Graph-Conditioned Feature Diffusion (GFD)}
\label{sec:CGD_ablations}

The generator of the main paper is trained on $\mathcal{L}_{\text{diff}}$ alone (Section~\ref{sec:cgd} of the main paper). This section documents the divergence-trained generator that serves as the ablation of Section~\ref{sec:controls} of the main paper (Section~\ref{subsec:divergence_generator}), its capacity ablation, and the configuration-selection procedure that fixed the architecture and the capacity, which the reconstruction-only generator inherits unchanged. Results are reported as frame-level AUC for UCF-Crime and Average Precision (AP) for XD-Violence.

\subsubsection{Models Reimplementation}
\begin{table}[!tb]
    \centering
    \footnotesize
    \caption{\textbf{Clean operating points of the detectors and the cost of regeneration.} GT is the score on the clean features in our evaluation, $\Delta_\text{graph}$ the drop from the clean features to the full-graph reconstruction of ETU at $t{=}300$, as in Table~\ref{tab:audit_final} of the main paper, and Publ. the value reported by the detector's own paper with the same features, where one exists. Detectors marked with {\color{red}$^*$} were trained by us. MIL is not audited; it is the detector used for configuration selection (Section~\ref{sec:CGD_ablations}).}
    \label{tab:vad_gt_graph_wrap}
    \setlength{\tabcolsep}{3pt}
    \fitcol{\begin{tabular*}{\linewidth}{@{\extracolsep{\fill}}l ccc ccc}
    \toprule
    \multirow{2}{*}{\textbf{Detector}} & \multicolumn{3}{c}{\textbf{UCF-Crime (AUC)}} & \multicolumn{3}{c}{\textbf{XD-Violence (AP)}} \\
    \cmidrule(lr){2-4} \cmidrule(lr){5-7}
     & \textbf{Publ.} & \textbf{GT} & $\Delta_\text{graph}$ & \textbf{Publ.} & \textbf{GT} & $\Delta_\text{graph}$\\
    \midrule
    \multicolumn{7}{@{}l}{\textbf{I3D features}} \\
    MIL{\color{red}$^*$} & -- & 83.49 & {2.37} & -- & 63.84 & {0.95} \\
    GlanceVAD & 91.96 & 91.76 & {1.66} & 89.40 & 89.18 & {0.20} \\
    MGFN{\color{red}$^*$} & \PublishedMgfnUcfIthreed & 80.80 & {3.25} & \PublishedMgfnXdvIthreed & 67.57 & {0.97} \\
    RTFM{\color{red}$^*$} & \PublishedRtfmUcfIthreed & 79.69 & {1.93} & \PublishedRtfmXdvIthreed & 74.03 & {0.70} \\
    UR-DMU & 86.97 & 86.97 & {1.70} & 81.66 & 81.66 & {0.22} \\
    PEL4VAD & 86.76 & 86.36 & {1.10} & 85.59 & 85.26 & {0.05} \\
    \midrule
    \multicolumn{7}{@{}l}{\textbf{CLIP features}} \\
    MIL{\color{red}$^*$} & -- & 84.26 & {0.44} & -- & 69.89 & {0.32} \\
    GlanceVAD{\color{red}$^*$} & -- & 87.39 & 0.30 & -- & 68.54 & {0.02} \\
    MGFN{\color{red}$^*$} & -- & 83.66 & 0.09 & -- & 72.33 & {1.24} \\
    RTFM{\color{red}$^*$} & -- & 85.45 & $-0.10$ & -- & 75.77 & {0.86} \\
    UR-DMU{\color{red}$^*$} & -- & 85.05 & 0.13 & -- & 81.72 & {0.71} \\
    DSANet & 89.44 & 89.44 & 0.16 & 86.95 & 86.95 & {0.65} \\
    VadCLIP & 88.02 & \VadClipAucHarness & 0.03 & 84.51 & 84.50 & {0.83} \\
    \bottomrule
    \end{tabular*}}
\end{table}
Our I3D re-implementations of MGFN and RTFM sit \ShortfallMgfnUcfIthreed{} and \ShortfallRtfmUcfIthreed{} AUC points below their published UCF-Crime operating points and \ShortfallMgfnXdvIthreed{} and \ShortfallRtfmXdvIthreed{} AP points below on XD-Violence, so no architectural reading is drawn from them in the main paper; the CLIP variants of MIL, GlanceVAD, MGFN, RTFM and UR-DMU were trained by us with the authors' code and have no published CLIP counterpart. 

\subsubsection{The Divergence-Trained Generator}
\label{subsec:divergence_generator}
The divergence-trained generator is the same architecture trained with $\mathcal{L}_{\text{GFD}} = \mathcal{L}_{\text{diff}} + \mathcal{L}^{\text{GFD}}_{\text{cont}}$, where, for each training video, one noise vector $\varepsilon$ and one timestep $t$ are shared by its four conditioning latents, $\hat{\mathbf{x}}_m$ is the estimate of $\mathbf{x}_0$ under $\mathbf{z}_m$, and $\mathcal{L}^{\text{GFD}}_{\text{cont}}$ keeps the irrelevant-mask estimate on the full-graph one and pushes the relevant- and semantic-mask estimates at least a margin away from it, with margins $\gamma_{\text{rel}}, \gamma_{\text{sem}} > 0$ and weights $\beta_{\text{irr}} = 0.65$, $\beta_{\text{rel}} = 0.60$ and $\beta_{\text{sem}} = 0.45$:
\begin{equation}
\label{eq:cgd_cont}
\begin{split}
    \mathcal{L}^{\text{GFD}}_{\text{cont}} = {} & \beta_\text{irr}\|\hat{\mathbf{x}}_{\text{graph}} - \hat{\mathbf{x}}_{\text{irr}}\|_2^2 \\
    & + \beta_\text{rel}\max(0, \gamma_{\text{rel}} - \|\hat{\mathbf{x}}_{\text{graph}} - \hat{\mathbf{x}}_{\text{rel}}\|_2^2) \\
    & + \beta_\text{sem}\max(0, \gamma_{\text{sem}} - \|\hat{\mathbf{x}}_{\text{graph}} - \hat{\mathbf{x}}_{\text{sem}}\|_2^2)
\end{split}
\end{equation}
Irrelevant perturbations are thereby trained to yield invariant reconstructions and outcome-chain interruptions divergent ones, so this generator is trained to produce the contrast that the audit measures. As Section~\ref{sec:controls} of the main paper shows, for four of the six detectors its deltas follow the displacement of the reconstruction, one untrained random node costs it 9.91 points on average against 15.18 for the whole outcome chain, and its count-matched deletion does not separate the relevant mask from the random draw (Table~\ref{tab:regimes} of the main paper). Table~\ref{tab:divgen_states} reports it on the full test set. Table~\ref{tab:ctrl_randnode} reports a retrain in which a conditioning variant with one random node removed is added to its training; afterwards one random node changes the AUC by at most \CtrlRandTrained{} points on every detector. This follows from training on that variant and is not evidence about other deletions; the relevant mask still costs between \RandkRelLo{} and \RandkRelHi{} points on it.

\begin{table}[!t]
    \centering
    \scriptsize
    \caption{\textbf{Random-node retrain of the divergence-trained generator.} The same generator, retrained with an added random-node conditioning variant, on all \RandCtrlNTotal{} UCF-Crime/CLIP test videos. Frame-AUC points lost from this model's own full-graph reconstruction under one random node, a conditioning variant this model saw in training, and under the relevant mask, which still moves every detector (bold).}
    \label{tab:ctrl_randnode}
    \setlength{\tabcolsep}{1.8pt}
    \fitcol{\begin{tabular*}{\linewidth}{@{\extracolsep{\fill}}@{}l rrrrrr@{}}
    \toprule
    \textbf{Setting} & MGFN & UR-DMU & RTFM & DSANet & VadCLIP & GlanceVAD \\
    \midrule
    $\Delta_{\text{rand}}$, one node & $-0.09$ & $0.00$ & $-0.02$ & $-0.02$ & $-0.04$ & $-0.01$ \\
$\Delta_{\text{rel}}$ & $\mathbf{1.03}$ & $\mathbf{1.19}$ & $\mathbf{5.01}$ & $\mathbf{16.50}$ & $\mathbf{6.90}$ & $\mathbf{1.01}$ \\

    \bottomrule
    \end{tabular*}}
\end{table}
\begin{table}[!t]
    \centering
    \scriptsize
    \caption{\textbf{The divergence-trained generator on the full test set}. Frame-AUC points lost from the full-graph reconstruction (Eq.~\ref{eq:delta} of the main paper); $\Delta_{\text{uncond}}$ is the cost of regenerating without any conditioning and $\Delta_{\text{rand}}$ (bold) that of one random node not seen in training (9.91 points on average, against 15.18 for the relevant mask). Its count-matched deletion is in Table~\ref{tab:regimes} of the main paper.}
    \label{tab:divgen_states}
    \setlength{\tabcolsep}{2.5pt}
    \fitcol{\begin{tabular*}{\linewidth}{@{\extracolsep{\fill}}@{}l r rrrr@{}}
    \toprule
    \textbf{Detector} & $\Delta_{\text{graph}}$ & $\Delta_{\text{uncond}}$ & $\Delta_{\text{irr}}$ & $\Delta_{\text{rand}}$ & $\Delta_{\text{rel}}$ \\
    \midrule
    MGFN & $0.24$ & $0.03$ & $0.04$ & $\mathbf{14.29}$ & $8.76$ \\
UR-DMU & $0.17$ & $-0.10$ & $-0.02$ & $\mathbf{7.85}$ & $-2.10$ \\
RTFM & $0.20$ & $-0.15$ & $-0.13$ & $\mathbf{14.92}$ & $-0.46$ \\
DSANet & $0.97$ & $-0.01$ & $0.07$ & $\mathbf{13.31}$ & $42.81$ \\
VadCLIP & $0.55$ & $0.05$ & $-0.01$ & $\mathbf{7.32}$ & $32.75$ \\
GlanceVAD & $0.49$ & $-0.04$ & $0.03$ & $\mathbf{1.75}$ & $9.31$ \\

    \bottomrule
    \end{tabular*}}
\end{table}

\begin{table}[!t]
    \centering
    \footnotesize
    \caption{\textbf{Dimension-level deletion at a matched 10\% budget on the CLIP track.} Frame-AUC change from deleting the 10\% of feature dimensions ranked highest by magnitude (Mag.), by input gradient (Grad.; four detectors, -- where unavailable) or at random (Rand., 5 trials), against the same run's clean score (GT, the baseline evaluation's own measurement, within 0.6 points of the UCF-Crime GT column of Table~\ref{tab:vad_gt_graph_wrap}; XD-Violence is reported here in AUC and there in AP). Both datasets are in frame-AUC here. The ranking is computed inside the audited detector, so the budget matches across detectors but the deleted content does not. The table gives context for the scale of the interventions and is not a method that they are compared against.}
    \label{tab:baselines}
    \setlength{\tabcolsep}{3pt}
    \fitcol{\begin{tabular}{@{}l cccc cccc}
    \toprule
    & \multicolumn{4}{c}{\textbf{UCF-Crime (AUC)}} & \multicolumn{4}{c}{\textbf{XD-Violence (AUC)}} \\
    \cmidrule(lr){2-5} \cmidrule(lr){6-9}
    \textbf{Detector} & GT & Mag. & Grad. & Rand. & GT & Mag. & Grad. & Rand. \\
    \midrule
    UR-DMU & $85.07$ & $-5.13$ & $-10.70$ & $0.68$ & $94.43$ & $-2.95$ & $-1.36$ & $0.31$ \\
RTFM & $85.45$ & $-15.37$ & $-25.04$ & $-0.18$ & $92.45$ & $-7.95$ & $-6.25$ & $0.71$ \\
MGFN & $83.65$ & $-4.12$ & $-14.38$ & $0.66$ & $91.48$ & $-4.97$ & $0.89$ & $-0.12$ \\
GlanceVAD & $87.93$ & $-3.50$ & $-3.26$ & $0.16$ & $95.30$ & $-2.28$ & $-1.93$ & $-0.07$ \\
DSANet & $88.97$ & $-5.41$ & -- & $-0.30$ & $94.84$ & $-8.08$ & -- & $-0.02$ \\
VadCLIP & $85.82$ & $-6.13$ & -- & $-0.16$ & $93.56$ & $-5.30$ & -- & $-0.48$ \\

    \bottomrule
    \end{tabular}}
\end{table}

Configurations are selected with a selectivity criterion that prioritizes intervention sensitivity over raw anomaly detection performance; the reconstruction-only generator inherits the selected architecture, capacity, curriculum and hyperparameters, with every weight of Eq.~\ref{eq:cgd_cont} set to zero, and every configuration is trained with {three} seeds. A configuration must first keep the cost of the full-graph reconstruction for the evaluating detector below $2$\% AUC, and is then ranked by its selectivity gap, the mean absolute score change under the relevant and semantic masks minus the mean under the full graph and the irrelevant mask, with lower leakage and then higher AUC as tie-breakers. 

\textbf{Selection split.} The criterion is computed on {a held-out subset of the training videos} with a MIL detector~\cite{sultani2018real} that is not among the detectors audited in the main paper, so the test videos never enter the selection and the audited detectors never enter the selection objective. The dimension-deletion baselines received no equivalent tuning.

\subsubsection{Architecture and Conditioning}

The denoiser employs a Transformer-based architecture formulated for noise prediction ($\epsilon$ prediction). At each diffusion timestep $t$, the network receives noisy video features $\mathbf{x}_t$ and a graph conditioning embedding $\mathbf{z}$, predicting the added noise $\epsilon$.

Graph conditioning is injected via two complementary pathways. First, Feature-wise Linear Modulation (FiLM)~\cite{perez2018film} utilizes a linear projection to map $\mathbf{z} \in \mathbb{R}^{512}$ to scale and shift parameters applied following each layer normalization. Second, a cross-attention mechanism projects $\mathbf{z}$ to learned conditioning tokens that function as keys and values. To accommodate varying feature extractors, a linear input projection compresses I3D features ($d_\text{feat} = 1024$) to the model dimension ($d_\text{model} = 512$) prior to processing, whereas CLIP features ($d_\text{feat} = 512$) natively match the internal dimension.

\textbf{Evaluated checkpoint.} All results use the final checkpoint of training. 

\subsubsection{Architecture Capacity}

To determine if the baseline model represents a capacity bottleneck for high-dimensional I3D inputs, we systematically increased the dimensions of the Transformer layers and cross-attention heads.

\begin{table}[ht]
    \centering
    \caption{Architecture capacity ablation of the divergence-trained generator (UCF-Crime / I3D); the reconstruction-only generator keeps the selected size.}
    \fitcol{\begin{tabular*}{\linewidth}{@{\extracolsep{\fill}}cccccc}
    \toprule
    $d_\text{model}$ & \textbf{FFN} & \textbf{Heads} & \textbf{Tokens} & \textbf{Params} & $\Delta$ \\
    \midrule
768 & 1536 & 8 & 4 & $\sim$38M & $-0.087$ \\
1024 & 2048 & 8 & 8 & $\sim$68M & $-0.088$ \\
    \rowcolor{gray!25} \textbf{512} & \textbf{1024} & \textbf{4} & \textbf{4} & \textbf{$\sim$7M} & \textbf{$-0.086$} \\
    \bottomrule
    \end{tabular*}}
    \label{tab:architecture_capacity}
\end{table}

The three deltas of Table~\ref{tab:architecture_capacity} differ by 0.002 in absolute value, so capacity had no measurable effect in this range, and we keep the smallest model. 

\textbf{GFD.} Tables~\ref{tab:cgd_ucf_wrap} and~\ref{tab:xdv_cgd_ablation} evaluate the GFD denoising process via cosine similarity and $MSE$ relative to ground-truth features. Cosine similarity exceeds $0.9$ under $\mathbf{z}_{\text{graph}}$ and $\mathbf{z}_{\text{irr}}$, confirming accurate directional reconstruction. %
\begin{table}[!tb]
    \centering
    \small
    \caption{GFD reconstruction on UCF-Crime. Cells are means over all \RandCtrlNTotal{} test videos; Table~\ref{tab:displacement_supp} reports medians over the test videos for both generators, including the count-matched random draw.}
    \label{tab:cgd_ucf_wrap}
    \setlength{\tabcolsep}{3pt}
    \fitcol{\begin{tabular*}{\linewidth}{@{\extracolsep{\fill}}ll | cc | cc}
        \toprule
        \textbf{Enc.} & \textbf{Cond.} & \multicolumn{2}{c|}{\textbf{vs. $\mathbf{x}_0$}} & \multicolumn{2}{c}{\textbf{vs. $\mathbf{\hat{x}}_{\text{graph}}$}} \\
        \cmidrule(lr){3-4} \cmidrule(lr){5-6}
        & & \textbf{Cos.} & \textbf{MSE} & \textbf{Cos.} & \textbf{MSE} \\
        \midrule
        \multirow{4}{*}{I3D} 
        & $\mathbf{z}_{g}$ & {0.956} & {0.008} & -- & -- \\
        & $\mathbf{z}_{\text{irr}}$ & {0.954} & {0.009} & {0.998} & {0.001} \\
        & $\mathbf{z}_{\text{rel}}$ & {0.887} & {0.043} & {0.931} & {0.053} \\
        & $\mathbf{z}_{\text{sem}}$ & {0.856} & {0.068} & {0.899} & {0.063} \\
        \midrule
        \multirow{4}{*}{CLIP} 
        & $\mathbf{z}_{g}$ & {0.978} & {0.008} & -- & -- \\
        & $\mathbf{z}_{\text{irr}}$ & {0.977} & {0.012} & {0.996} & {0.003} \\
        & $\mathbf{z}_{\text{rel}}$ & {0.687} & {0.251} & {0.708} & {0.238} \\
        & $\mathbf{z}_{\text{sem}}$ & {0.218} & {1.330} & {0.232} & {1.311} \\
        \bottomrule
    \end{tabular*}}
\end{table}

\begin{table}[ht!]
    \caption{GFD reconstruction on XD-Violence.}
    \label{tab:xdv_cgd_ablation}
    \centering
    \fitcol{\begin{tabular}{@{}ll cccccc}
        \toprule
        \multirow{2}{*}{\textbf{Enc.}} & \multirow{2}{*}{\textbf{Cond.}} & \multicolumn{3}{c}{\textbf{vs. Ground-Truth ($\mathbf{x}_0$)}} & \multicolumn{3}{c}{\textbf{vs. Full Graph ($\mathbf{x}_{\text{graph}}$)}} \\
        \cmidrule(lr){3-5} \cmidrule(lr){6-8}
        & & \textbf{Cos. Sim.} & \textbf{MSE} & & \textbf{Cos. Sim.} & \textbf{MSE} & \\
        \midrule
        \multirow{4}{*}{I3D} 
        & $\mathbf{z}_{\text{graph}}$ & {0.942 $\uparrow$} & {0.010 $\downarrow$} & & -- & -- & \\
        & $\mathbf{z}_{\text{irr}}$   & {0.940 $\uparrow$} & {0.011 $\downarrow$} & & 0.997 $\uparrow$ & 0.001 $\downarrow$ & \\
        & $\mathbf{z}_{\text{rel}}$   & {0.903 $\downarrow$} & {0.020 $\uparrow$} & & 0.960 $\downarrow$ & 0.010 $\uparrow$ & \\
        & $\mathbf{z}_{\text{sem}}$   & {0.500 $\downarrow$} & {0.315 $\uparrow$} & & 0.536 $\downarrow$ & 0.304 $\uparrow$ & \\
        \midrule
        \multirow{4}{*}{CLIP} 
        & $\mathbf{z}_{\text{graph}}$ & {0.966 $\uparrow$} & {0.015 $\downarrow$} & & -- & -- & \\
        & $\mathbf{z}_{\text{irr}}$   & {0.965 $\uparrow$} & {0.016 $\downarrow$} & & 0.998 $\uparrow$ & 0.001 $\downarrow$ & \\
        & $\mathbf{z}_{\text{rel}}$   & {0.712 $\downarrow$} & {0.229 $\uparrow$} & & 0.751 $\downarrow$ & 0.206 $\uparrow$ & \\
        & $\mathbf{z}_{\text{sem}}$   & {0.386 $\downarrow$} & {0.595 $\uparrow$} & & 0.427 $\downarrow$ & 0.559 $\uparrow$ & \\
        \bottomrule
    \end{tabular}}
    
\end{table}

\section{VLM-based Counterfactual Explainability}
\label{sec:vlm_pitfall}
\begin{figure}[!t]
    \centering
    \includegraphics[width=\linewidth]{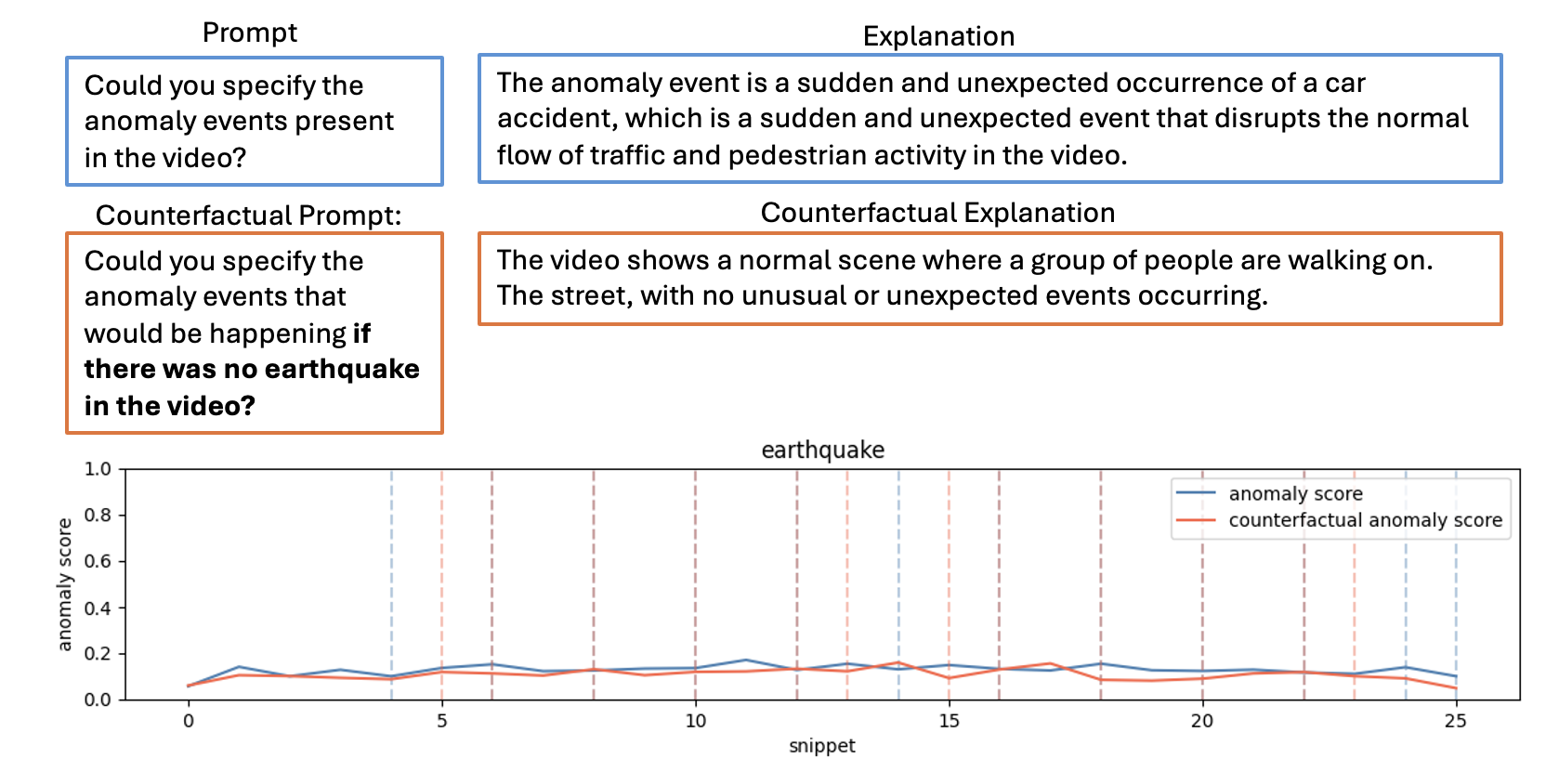}
    \caption{Anomaly scores and explanation of ``earthquake.mp4'' from the official Holmes-VAU repository~\cite{zhang2025holmes}. The top panel shows the default and the counterfactual prompt with the explanation returned for each, and the bottom panel the per-snippet anomaly scores under each prompt.}
    \label{fig:holmes_earthquake}
\end{figure}
\begin{figure}[!t]
    \centering
    \includegraphics[width=\linewidth]{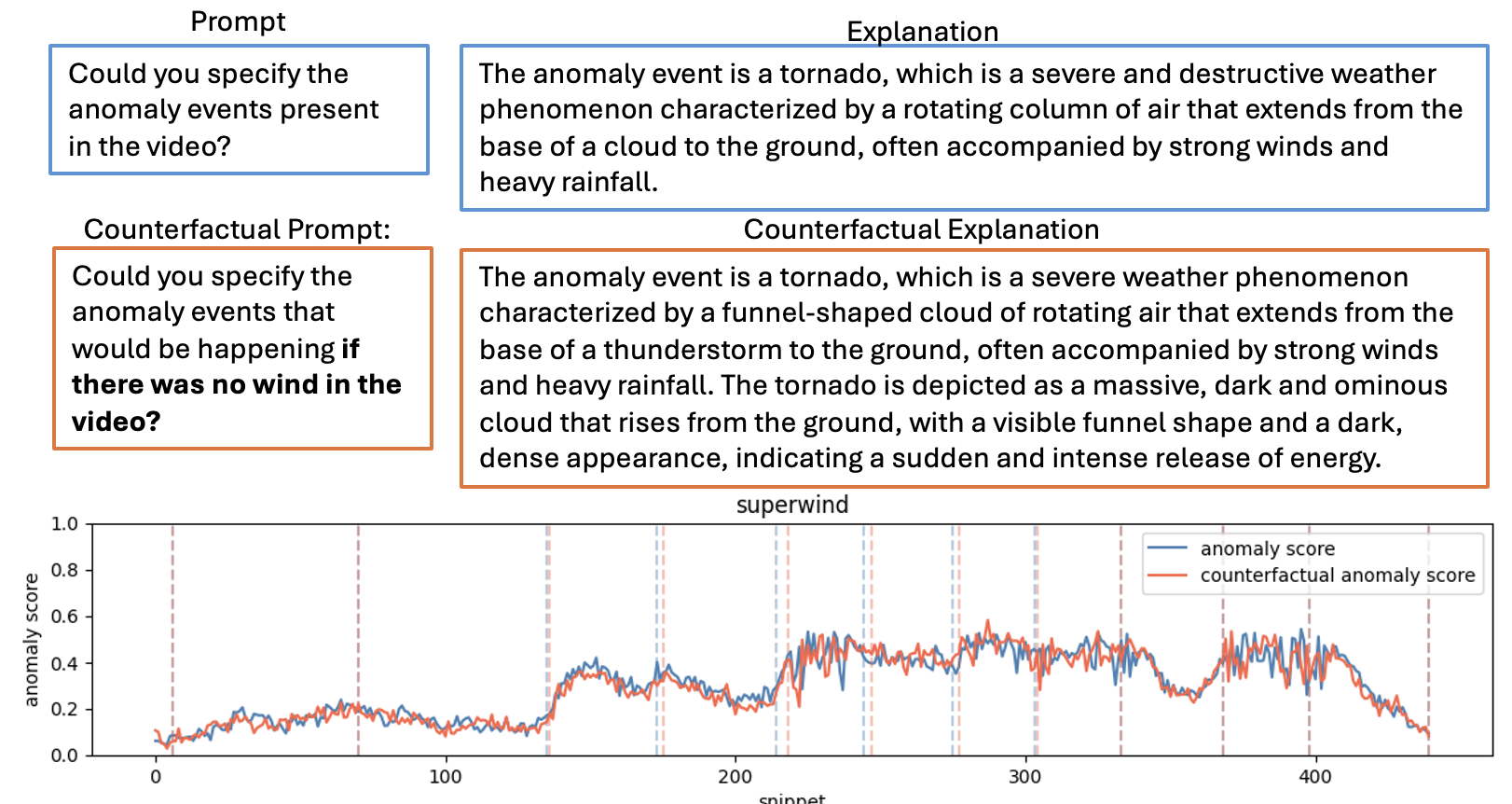}
    \caption{Anomaly scores and explanation of ``superwind.mp4'' from the official Holmes-VAU repository~\cite{zhang2025holmes}. The top panel shows the default and the counterfactual prompt with the explanation returned for each, and the bottom panel the per-snippet anomaly scores under each prompt.}
    \label{fig:holmes_superwind}
\end{figure}
This section adds two videos from the official repository of Holmes-VAU~\cite{zhang2025holmes} to the ``robbery'' video of Figure~\ref{fig:holmes1} of the main paper. For each video we pose the default prompt and a counterfactual prompt that asks which anomalies would occur without a named element of the scene, and plot the per-snippet anomaly scores under the two prompts.

In the ``earthquake'' video (Figure~\ref{fig:holmes_earthquake}), the explanation under the default prompt names ``a sudden and unexpected occurrence of a car accident'' as the anomaly; under the counterfactual prompt (no earthquake) it describes a normal street scene with no unusual events. Both score curves stay below 0.2 on every snippet, so the default explanation reports an anomaly that the score does not register. In the ``superwind'' video (Figure~\ref{fig:holmes_superwind}), the counterfactual prompt asks which anomalies would occur without the wind; the explanation names the same anomaly as under the default prompt, a tornado, ``[...] often accompanied by strong winds [...]'', so it ignores the counterfactual premise, and the two score curves are again nearly identical. Across the three videos the score is nearly unchanged under the counterfactual prompt whether the explanation changes or not.

\section{Interval Estimates and Displacement for the Count-Matched Deletion}
\label{sec:supp_intervals}
Every confidence interval that this paper reports for the count-matched random deletion comes from one procedure. The resampling unit is the video, and each replicate is one draw with replacement of the UCF-Crime test videos of Table~\ref{tab:regimes} of the main paper, applied in lockstep to the full-graph, relevant-mask and random-deletion conditionings and to all six detectors, so that every quantity compared here is paired on the same videos. We draw $B = \BootReplicates$ replicates, recompute the benchmark metric inside each replicate (frame AUC over the pooled frames of the replicate, not an average of per-video values), and report the two-sided percentile interval between the \BootPercentileLo{}th and the \BootPercentileHi{}th percentiles of the replicate distribution. The intervals resample videos only. The difference $\Delta_{\text{rel}} - \Delta_{\text{rand}}$ is in frame-AUC points. The ratio $|\Delta_{\text{rand}}|/\Delta_{\text{rel}}$ is dimensionless and undefined on a replicate whose $\Delta_{\text{rel}}$ is not positive, so those replicates are dropped from the interval of the ratio only (at most \BootRatioGuardMax{} of $\BootReplicates$ for any detector at $t{=}300$).

Two-sided empirical $p$-values for a non-zero difference follow from the same replicate distribution. Under a Holm step-down at family-wise level \HolmFamilyAlpha, all six detectors reject; the largest $p$-value is that of VadCLIP, and the Holm threshold at its rank is \HolmFamilyAlpha{} itself.

Table~\ref{tab:intervals} reports both ends of every interval in blocks (i) and (iv) and the lower end in block (iii). Blocks (ii) and (iii) split the test videos by label. Block (ii) is the pooled frame AUC over the anomalous videos of Table~\ref{tab:regimes} of the main paper alone, as point estimates, since no resampling was run for that subset. Block (iii) is the paired mean shift of the raw detector score on the normal videos of Table~\ref{tab:regimes} of the main paper, {where the count-matched draw displaces the reconstruction farther (median MSE 0.051 against 0.009, Table~\ref{tab:displacement_supp}) and moves the scores \SplitNormalShiftRatioLo{} to \SplitNormalShiftRatioHi{} times less}. Block (iv) is the divergence-free ETU.
\begin{table}[!tb]
    \centering
    \scriptsize
    \caption{\textbf{Confidence intervals for the count-matched deletion.} UCF-Crime with CLIP features at $t{=}300$ on ETU, except block (iv), the divergence-free ETU. Blocks are different video sets or quantities and do not compare across a block boundary; block (iii) is in raw score units, since pooled AUC is undefined on videos without positive frames. The differences of block (i) are in bold.}
    \label{tab:intervals}
    \setlength{\tabcolsep}{1.5pt}
    \fitcol{\begin{tabular*}{\linewidth}{@{\extracolsep{\fill}}@{}l rrrrrr@{}}
    \toprule
    \textbf{Quantity} & MGFN & UR-DMU & RTFM & DSANet & VadCLIP & GlanceVAD \\
    \midrule
    \multicolumn{7}{@{}l}{\textit{(i) ETU, test set videos, frame-AUC points}} \\[1pt]
$\Delta_{\text{rel}} - \Delta_{\text{rand}}$ & {$\mathbf{8.23}$} & {$\mathbf{5.89}$} & {$\mathbf{5.91}$} & {$\mathbf{5.45}$} & {$\mathbf{5.34}$} & {$\mathbf{4.72}$} \\
\quad 95\% lower & {$4.89$} & {$3.98$} & {$2.93$} & {$3.22$} & {$0.47$} & {$4.45$} \\
\quad 95\% upper & {$21.36$} & {$15.27$} & {$18.67$} & {$14.21$} & {$22.76$} & {$19.23$} \\
\quad two-sided $p$ & {$0.0002$} & {$0.0020$} & {$0.0032$} & {$0.0008$} & {$0.0342$} & {$<0.0002$} \\
$|\Delta_{\text{rand}}|/\Delta_{\text{rel}}$ & {$0.214$} & {$0.131$} & {$0.127$} & {$0.288$} & {$0.175$} & {$0.073$} \\
\quad 95\% lower & {$0.006$} & {$0.005$} & {$0.005$} & {$0.008$} & {$0.004$} & {$0.003$} \\
\quad 95\% upper & {$0.526$} & {$0.506$} & {$0.497$} & {$0.829$} & {$0.843$} & {$0.300$} \\
\midrule
\multicolumn{7}{@{}l}{\textit{(ii) ETU, anomalous videos, point estimates}} \\[1pt]
$\Delta_{\text{rel}}$ & {$7.13$} & {$8.71$} & {$11.43$} & {$14.63$} & {$9.36$} & {$11.49$} \\
$\Delta_{\text{rand}}$ & {$-4.65$} & {$-1.67$} & {$1.90$} & {$-2.34$} & {$3.18$} & {$1.02$} \\
difference & {$11.77$} & {$10.39$} & {$9.53$} & {$16.97$} & {$6.18$} & {$10.47$} \\
\midrule
\multicolumn{7}{@{}l}{\textit{(iii) ETU, normal videos, raw score shift}} \\[1pt]
relevant mask & {$0.324$} & {$0.148$} & {$0.080$} & {$0.098$} & {$0.189$} & {$0.116$} \\
\quad 95\% lower & {$0.240$} & {$0.089$} & {$0.039$} & {$0.076$} & {$0.125$} & {$0.079$} \\
random draw & {$0.068$} & {$0.048$} & {$0.034$} & {$0.021$} & {$0.055$} & {$0.020$} \\
\quad 95\% lower & {$0.006$} & {$0.013$} & {$0.016$} & {$-0.001$} & {$-0.001$} & {$-0.003$} \\
\midrule
\multicolumn{7}{@{}l}{\textit{(iv) Divergence-free ETU, videos of Table 3 of the main paper, frame-AUC points}} \\[1pt]
$\Delta_{\text{rel}}$ & {$\DivFreeRelMgfn$} & {$\DivFreeRelUrdmu$} & {$\DivFreeRelRtfm$} & {$\DivFreeRelDsanet$} & {$\DivFreeRelVadclip$} & {$\DivFreeRelGlancevad$} \\
$\Delta_{\text{rand}}$ & {$\DivFreeRandMgfn$} & {$\DivFreeRandUrdmu$} & {$\DivFreeRandRtfm$} & {$\DivFreeRandDsanet$} & {$\DivFreeRandVadclip$} & {$\DivFreeRandGlancevad$} \\
$\Delta_{\text{rel}} - \Delta_{\text{rand}}$ & {$\DivFreeDiffMgfn$} & {$\DivFreeDiffUrdmu$} & {$\DivFreeDiffRtfm$} & {$\DivFreeDiffDsanet$} & {$\DivFreeDiffVadclip$} & {$\DivFreeDiffGlancevad$} \\
\quad 95\% lower & {$\DivFreeCiLoMgfn$} & {$\DivFreeCiLoUrdmu$} & {$\DivFreeCiLoRtfm$} & {$\DivFreeCiLoDsanet$} & {$\DivFreeCiLoVadclip$} & {$\DivFreeCiLoGlancevad$} \\
\quad 95\% upper & {$\DivFreeCiHiMgfn$} & {$\DivFreeCiHiUrdmu$} & {$\DivFreeCiHiRtfm$} & {$\DivFreeCiHiDsanet$} & {$\DivFreeCiHiVadclip$} & {$\DivFreeCiHiGlancevad$} \\

    \bottomrule
    \end{tabular*}}
\end{table}

Table~\ref{tab:noise_sweep} reports the three masks of ETU at the three noise levels. The irrelevant mask stays within \FidelityIrrCap{} at every level, and the relevant mask grows monotonically with $t$. Tables~\ref{tab:displacement_supp} and~\ref{tab:latent_displacement} report the displacement behind the intervals of Table~\ref{tab:intervals}, at the level of the reconstruction and of the conditioning latent. {On ETU the count-matched random draw displaces the reconstruction at least as far as the relevant mask on the normal videos and about as far on the anomalous ones,} while on the divergence-trained generator the relevant-mask reconstruction of the anomalous videos is nearly orthogonal to the full-graph one (median cosine 0.051) and the random draw moves farther still ($-0.217$). {At the level of the latent, the random latent is the farther of the two in cosine on \LatRandFartherPct\% of the videos, and on the anomalous videos the two latents are within \LatCosAnomGap{} of one another ($\LatCosRandAnom$ against $\LatCosRelAnom$).} The asymmetry between the relevant and the random deletion on ETU is therefore not inherited from the geometry that the contrastive objective of Eq.~\ref{eq:gem_cont} trains. {This holds for the cosine only. In $L_2$ the relevant latent moves farther on every video set ($\LatLtwoRelAnom$ against $\LatLtwoRandAnom$), since the hinge of Eq.~\ref{eq:gem_cont} constrains the direction and not the magnitude, and ETU does not pass that difference on (Table~\ref{tab:displacement_supp}).}

\begin{table}[!tb]
    \centering
    \scriptsize
    \caption{\textbf{Reconstruction displacement of the three masks, split by video label.} Cosine and MSE of each masked reconstruction to the full-graph reconstruction of the same run, per-video mean over segments and median over the videos of Table~\ref{tab:regimes} of the main paper, on ETU and on the divergence-trained generator. The last row gives the per-video median ratio random$/$relevant of $1-\cos$ and of MSE. The larger displacement on ETU over all videos is in bold.}
    \label{tab:displacement_supp}
    \setlength{\tabcolsep}{3pt}
    \fitcol{\begin{tabular*}{\linewidth}{@{\extracolsep{\fill}}@{}l rr rr@{}}
    \toprule
    & \multicolumn{2}{c}{\textbf{ETU}} & \multicolumn{2}{c}{\textbf{Divergence-trained}} \\
    \cmidrule(lr){2-3} \cmidrule(lr){4-5}
    \textbf{Mask} & $\cos$ & MSE & $\cos$ & MSE \\
    \midrule
    \multicolumn{5}{@{}l}{\textit{videos of Table 3 of the main paper}} \\[1pt]
irrelevant mask & {$1.000$} & {$0.000$} & {$1.000$} & {$0.000$} \\
relevant mask & {$0.977$} & {$0.009$} & {$0.999$} & {$0.000$} \\
count-matched random & {$\mathbf{0.955}$} & {$\mathbf{0.020}$} & {$-0.003$} & {$0.895$} \\
\midrule
\multicolumn{5}{@{}l}{\textit{anomalous videos}} \\[1pt]
irrelevant mask & {$1.000$} & {$0.000$} & {$1.000$} & {$0.000$} \\
relevant mask & {$0.983$} & {$0.006$} & {$0.051$} & {$0.553$} \\
count-matched random & {$0.989$} & {$0.005$} & {$-0.217$} & {$1.715$} \\
\midrule
\multicolumn{5}{@{}l}{\textit{normal videos}} \\[1pt]
irrelevant mask & {$1.000$} & {$0.000$} & {$1.000$} & {$0.000$} \\
relevant mask & {$0.976$} & {$0.009$} & {$1.000$} & {$0.000$} \\
count-matched random & {$0.939$} & {$0.051$} & {$0.141$} & {$0.600$} \\
\midrule
random$/$relevant ratio, per video & {$0.95$} & {$1.11$} & {$2.41$} & {$5.44$} \\

    \bottomrule
    \end{tabular*}}
\end{table}
\begin{table}[!tb]
    \centering
    \scriptsize
    \caption{\textbf{In cosine, the conditioning latent moves at least as far under the count-matched random draw as under the relevant mask on all and on normal videos; in $L_2$ the relevant latent moves farther.} Median cosine and $L_2$ distance of each masked GEM latent to $\mathbf{z}_{\text{graph}}$ on the videos of Table~\ref{tab:displacement_supp}; ETU and the divergence-trained generator condition on this embedding. Bold marks the cells in which the random latent lies farther from $\mathbf{z}_{\text{graph}}$ than the relevant one.}
    \label{tab:latent_displacement}
    \setlength{\tabcolsep}{3pt}
    \fitcol{\begin{tabular*}{\linewidth}{@{\extracolsep{\fill}}@{}l rrr rrr@{}}
    \toprule
    & \multicolumn{3}{c}{\textbf{Cosine to} $\mathbf{z}_{\text{graph}}$} & \multicolumn{3}{c}{$\boldsymbol{L_2}$ \textbf{to} $\mathbf{z}_{\text{graph}}$} \\
    \cmidrule(lr){2-4} \cmidrule(lr){5-7}
    \textbf{Mask} & all & anom. & normal & all & anom. & normal \\
    \midrule
    irrelevant mask & {$0.963$} & {$0.948$} & {$0.984$} & {$0.9$} & {$1.2$} & {$0.7$} \\
relevant mask & {$0.458$} & {$-0.331$} & {$0.961$} & {$20.7$} & {$14.3$} & {$21.6$} \\
count-matched random & {$\mathbf{-0.235}$} & {$-0.307$} & {$\mathbf{-0.157}$} & {$12.1$} & {$9.8$} & {$13.0$} \\

    \bottomrule
    \end{tabular*}}
\end{table}
\begin{table}[!tb]
    \centering
    \scriptsize
    \caption{\textbf{Noise-level sweep of ETU} on all \RandCtrlNTotal{} UCF-Crime/CLIP test videos. Frame-AUC points lost from the full-graph reconstruction under each mask when the regeneration starts at noise level $t$; the default noise level, $t{=}300$, is in bold for $\Delta_{\text{rel}}$ and $\Delta_{\text{sem}}$.}
    \label{tab:noise_sweep}
    \setlength{\tabcolsep}{1.6pt}
    \fitcol{\begin{tabular*}{\linewidth}{@{\extracolsep{\fill}}@{}l rrr rrr rrr@{}}
    \toprule
    & \multicolumn{3}{c}{$\Delta_{\text{irr}}$ at $t$} & \multicolumn{3}{c}{$\Delta_{\text{rel}}$ at $t$} & \multicolumn{3}{c}{$\Delta_{\text{sem}}$ at $t$} \\
    \cmidrule(lr){2-4} \cmidrule(lr){5-7} \cmidrule(lr){8-10}
    \textbf{Detector} & 100 & 300 & 600 & 100 & 300 & 600 & 100 & 300 & 600 \\
    \midrule
    MGFN & $-0.03$ & $-0.01$ & $-0.09$ & $1.09$ & $\mathbf{6.78}$ & $13.71$ & $4.26$ & {$\mathbf{37.49}$} & $38.73$ \\
UR-DMU & $0.00$ & $-0.02$ & $-0.02$ & $0.61$ & $\mathbf{5.21}$ & $14.49$ & $0.43$ & {$\mathbf{4.03}$} & $6.43$ \\
RTFM & $0.00$ & $0.00$ & $0.02$ & $0.55$ & $\mathbf{6.77}$ & $19.41$ & $3.72$ & {$\mathbf{16.48}$} & $22.64$ \\
DSANet & $0.00$ & $-0.01$ & $0.01$ & $0.40$ & $\mathbf{4.23}$ & $10.79$ & $4.92$ & {$\mathbf{37.52}$} & $37.99$ \\
VadCLIP & $-0.01$ & $-0.03$ & $-0.04$ & $0.61$ & $\mathbf{6.47}$ & $13.84$ & $3.77$ & {$\mathbf{22.58}$} & $25.84$ \\
GlanceVAD & $0.00$ & $-0.01$ & $-0.01$ & $0.36$ & $\mathbf{5.09}$ & $12.07$ & $5.12$ & {$\mathbf{31.75}$} & $35.54$ \\

    \bottomrule
    \end{tabular*}}
\end{table}

\section{Three-Teacher Ablation}
\label{sec:supp_teacher}
The UCF-Crime videos were parsed by three VLM teachers, Qwen3-VL (the teacher used throughout the main paper), InternVideo3 and Gemma-4, with the prompts of Figure~\ref{fig:teacher_prompts}. 

The models are prompted with the video and three prompts to obtain the nodes and edges of the three layers of the semantic graph $\mathcal{G}_x$ that represents video $x$. While we use a prompt for each layer, they all share the system prompt of Figure~\ref{fig:teacher_prompts}.

\begin{figure}[t]
\begin{promptbox}{Shared rules (system prompt)}
Frames are sampled at 1\,fps and marked with their time in seconds.
(1)~Describe only what is visible; no intentions, emotions or identities.
(2)~Never name event categories (e.g.\ \emph{robbery}, \emph{assault}, \emph{accident}, \emph{anomaly}); describe the physical event (``person hits person'').
(3)~Do not assume anything unusual happens.
(4)~Labels: lowercase words, the same words for the same thing.
(5)~Output JSON only.
\end{promptbox}

\begin{promptbox}{Layer 1: roots}
List every agent: people, groups, animals, moving vehicles, fire, smoke, water. Objects that are only acted upon are not agents. Label each with a category noun (``person'', ``car''). At most 25.
\promptout{\{roots: [\{id: R1, label: person, first\_seen: 0, last\_seen: 42\}]\}}
\end{promptbox}

\begin{promptbox}{Layer 2: actions}
Agents: $\langle$layer 1$\rangle$. List every action of these agents, ordinary or unusual: one verb with its object (``opens door''). Link each action to the agents that perform it. At most 25.
\promptout{\{actions: [\{id: A1, label: opens door, agents: [R1], start: 3, end: 5\}]\}}
\end{promptbox}

\begin{promptbox}{Layer 3: outcomes}
Agents: $\langle$layer 1$\rangle$. Actions: $\langle$layer 2$\rangle$. List every visible change of state caused by these actions, written as entity and new state (``window broken'', ``person lying on ground''), ordinary or unusual. Link each outcome to the actions that cause it. At most 25.
\promptout{\{outcomes: [\{id: O1, label: window broken, caused\_by: [A4], first\_visible: 17\}]\}}
\end{promptbox}
\caption{\textbf{Teacher prompts.} The three layers are built by three calls on the same video. Each call receives the output of the previous layers, so the edges are the ID references in \texttt{agents} and \texttt{caused\_by}.}
\label{fig:teacher_prompts}
\end{figure}

Table~\ref{tab:teacher_agreement} measures how far the graphs of two teachers on the same video agree. The atoms agree above both a same-class null and a random-video null, while the edges agree little, with a same-video edge $F_1$ of \TeacherEdgeFoneLo--\TeacherEdgeFoneHi{}, only 0.024 to 0.083 above the same-class null. We therefore treat the recovered topology as a conditioning structure and not as the structure of the scene.

\begin{table}[!tb]
    \centering
    \footnotesize
    \caption{\textbf{Agreement between the graphs of the three teachers}. Atom cosines are matched-pair medians per node type (root / action / outcome); the same-class null pairs a video with another video of its own anomaly class, the random-video null with any other video. Edge $F_1$ compares the two teachers' recovered edge sets on the same video. The paired win is the share of videos on which the same-video atom cosine exceeds the same-class null.}
    \label{tab:teacher_agreement}
    \setlength{\tabcolsep}{3pt}
    \fitcol{\begin{tabular}{@{}l ccc}
    \toprule
    & \textbf{Qwen3-VL /} & \textbf{Qwen3-VL /} & \textbf{InternVideo3 /} \\
    & \textbf{InternVideo3} & \textbf{Gemma-4} & \textbf{Gemma-4} \\
    \midrule
    matched atom cosine, same video & 0.780 / 0.793 / 0.772 & 0.764 / 0.777 / 0.728 & 0.766 / 0.779 / 0.737 \\
\quad same-class null & 0.704 / 0.729 / 0.715 & 0.705 / 0.722 / 0.687 & 0.709 / 0.731 / 0.693 \\
\quad random-video null & 0.668 / 0.683 / 0.667 & 0.674 / 0.687 / 0.658 & 0.682 / 0.702 / 0.669 \\
\midrule
paired win vs same-class null & 81.4\% & 74.1\% & 73.4\% \\
\midrule
edge $F_1$, same video & 0.250 & 0.167 & 0.138 \\
\quad same-class null & 0.167 & 0.143 & 0.111 \\

    \bottomrule
    \end{tabular}}
\end{table}

\section{Single-Node Masks}
\label{sec:supp_single_node}
Table~\ref{tab:single_node} reports the single-node masks of Section~\ref{sec:controls} of the main paper. On ETU, one masked node changes the frame-AUC by less than the smallest relevant-mask delta of Table~\ref{tab:ctrl_reconly} of the main paper at $t{=}300$ (\FidelityGateTthree{} points). An on-chain node raises the AUC of all six detectors, by 0.04 to \FidelityKoneCap{} points, and an off-chain node raises it for five detectors, by up to \FidelityKoneOffCap{} points (RTFM), and lowers it by 0.20 points for VadCLIP. On the divergence-trained generator, by contrast, the on-chain node costs UR-DMU 20.30 points against 1.12 for the off-chain one, DSANet the reverse (1.77 against 23.97), and MGFN, DSANet, VadCLIP and GlanceVAD lose more under the off-chain node, although the two masks displace the reconstruction by the same median amount (median per-video ratio of off-chain to on-chain displacement \DispRLtwo{} in $L_2$ and \DispRCosine{} in cosine distance).

\begin{table}[!tb]
    \centering
    \scriptsize
    \caption{\textbf{Single-node masks ($k{=}1$).} Frame-AUC points lost from each model's own full-graph reconstruction under one masked node on or off the outcome chain, on all \RandCtrlNTotal{} UCF-Crime/CLIP test videos. A negative value means that the AUC rises.}
    \label{tab:single_node}
    \setlength{\tabcolsep}{1.8pt}
    \fitcol{\begin{tabular*}{\linewidth}{@{\extracolsep{\fill}}@{}l rrrrrr@{}}
    \toprule
    \textbf{Setting} & MGFN & UR-DMU & RTFM & DSANet & VadCLIP & GlanceVAD \\
    \midrule
    \multicolumn{7}{@{}l}{\textbf{ETU}, all \RandCtrlNTotal{} test videos} \\[1pt]
on-chain node & {$-0.28$} & {$-0.06$} & {$-0.04$} & {$-0.07$} & {$-0.22$} & {$-0.28$} \\
off-chain node & {$-0.21$} & {$-0.98$} & {$-1.80$} & {$-0.22$} & {$0.20$} & {$-0.34$} \\
\midrule
\multicolumn{7}{@{}l}{\textbf{Divergence-trained generator}, its own single-node cache, the same masks} \\[1pt]
on-chain node & $0.73$ & $20.30$ & $34.02$ & $1.77$ & $1.37$ & $-1.99$ \\
off-chain node & $24.24$ & $1.12$ & $0.24$ & $23.97$ & $16.73$ & $0.89$ \\

    \bottomrule
    \end{tabular*}}
\end{table}

\section{Qualitative Results}
\label{sec:supp_qualitative}

\begin{table*}[t]
    \centering
    \scriptsize
    \caption{Per-concept deltas ($\Delta$, on the 0--1 scale of the per-video evaluation; negative = a drop, the opposite sign of Eq.~\ref{eq:delta} of the main paper) from masking one named node at a time on two UCF-Crime test videos (CLIP features) for MIL ($M$), UR-DMU ($U$) and RTFM ($R$); not Eq.~\ref{eq:delta}'s test-set benchmark-point delta. The upper three rows of each video are nodes on an outcome path, the lower three are off every outcome path. An illustration of the measurement on single videos.}
    \label{tab:concept_deltas_side_by_side}
    
    \hfill
    \begin{minipage}{0.32\textwidth}
        \centering
        \text{RoadAccidents002} \\
        \setlength{\tabcolsep}{2pt}
        \begin{tabular}{llrrr}
            \toprule
            \textbf{Type} & \textbf{Concept} & $\Delta_M$ & $\Delta_U$ & $\Delta_R$ \\
            \midrule
            Action & Driving & $-0.384$ & $-0.175$ & $-0.219$ \\
            Root   & Car 1 & $-0.392$ & $-0.134$ & $-0.200$ \\
            Outcome   & Car 1 crash & $-0.314$ & $-0.192$ & $-0.178$ \\
            \midrule
            Root    & Pers. 2 & $-0.004$ & $-0.001$ & 0.092 \\
            Outcome & Pers. 2 exit & 0.002 & $-0.008$ & 0.003 \\
            Root    & Person 1 & 0.001 & $-0.012$ & $-0.075$ \\
            \bottomrule
        \end{tabular}
    \end{minipage}
    \hfill
    \begin{minipage}{0.32\textwidth}
        \centering
        \text{Shoplifting027} \\
        \setlength{\tabcolsep}{2pt}
        \begin{tabular}{llrrr}
            \toprule
            \textbf{Type} & \textbf{Concept} & $\Delta_M$ & $\Delta_U$ & $\Delta_R$ \\
            \midrule
            Root & Pers. 1 & $-0.342$ & $-0.156$ & $-0.213$\\
            Action   & Takes & $-0.492$ & $-0.142$ & $-0.245$ \\
            Root   & Pers. 2 & $-0.311$ & $-0.195$ & $-0.193$ \\
            \midrule
            Outcome & Pers. 1 exit & 0.003 & 0.004 & $-0.003$ \\
            Root    & Cashier & 0.012 & $-0.013$ & $-0.012$ \\
            Action    & Walks & $-0.003$ & $-0.026$ & $-0.023$ \\
            \bottomrule
        \end{tabular}
    \end{minipage}
    \hfill\null
\end{table*}

\begin{figure}
    \centering
    \vspace{-.5cm}
    \includegraphics[width=1\linewidth]{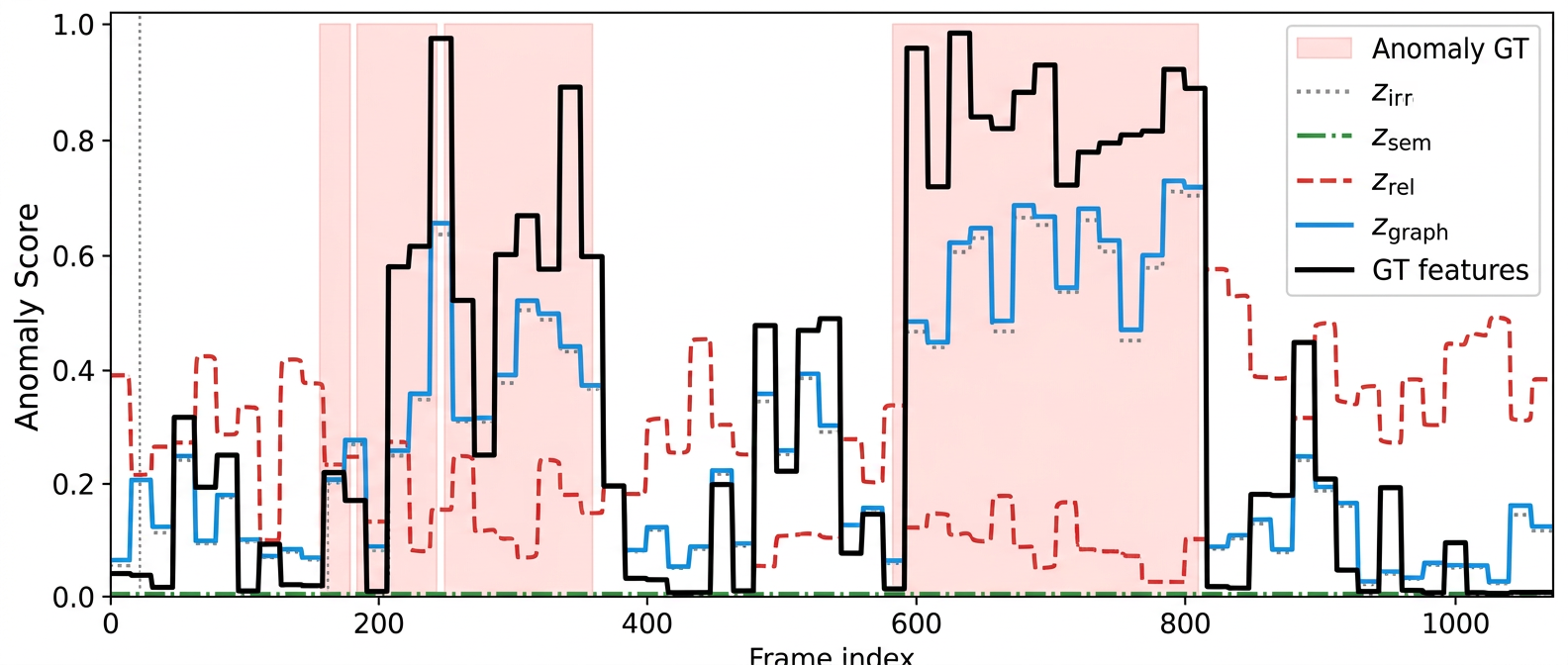}
    \caption{MIL anomaly-score traces on the \textit{Bad.Boys.1995-01-11-55-01-12-40} video of XD-Violence (CLIP features) under the ground-truth features and the four graph conditionings. In the legend, blue solid is $\mathbf{z_{graph}}$, grey dotted $\mathbf{z_{irr}}$, red dashed $\mathbf{z_{rel}}$, green dash-dotted $\mathbf{z_{sem}}$ and black solid the ground-truth features. Removing the outcome chain ($\mathbf{z_{rel}}$) or substituting the scene semantics ($\mathbf{z_{sem}}$) lowers the scores of this anomalous video; pruning off-path nodes ($\mathbf{z_{irr}}$) does not.}
    \label{fig:qualitative}
    \vspace{-.7cm}
\end{figure}

The figures of this section are per-video score traces of the MIL detector of Sultani et al.~\cite{sultani2018real} on XD-Violence with CLIP features, under the three graph masks and under the masking of single named nodes. Table~\ref{tab:concept_deltas_side_by_side} reports single-node masks on two UCF-Crime test videos. MIL is not among the detectors audited in the main paper; it is the detector on which the configuration-selection criterion was computed (Section~\ref{sec:CGD_ablations}). Every quantity here is a per-video curve or a per-video AP, not the test-set delta of Eq.~\ref{eq:delta} of the main paper. The traces illustrate what the measurement of the framework looks like on one video. They are not evidence that a named concept was removed from what the detector reads, and per-video AP is not monotone in the score level, so a mask can lower every score and raise the AP, or the reverse.

{Figure~\ref{fig:qualitative} shows the four graph conditionings on one video. Semantic masking ($\mathbf{z_{sem}}$) drives every score below $0.05$, relevant masking ($\mathbf{z_{rel}}$) lowers the scores inside the annotated windows against both the ground-truth features and the full-graph reconstruction ($\mathbf{z_{graph}}$), and pruning off-path nodes ($\mathbf{z_{irr}}$) leaves the trace nearly identical to $\mathbf{z_{graph}}$ and close to the ground-truth curve. This is the same ordering that Table~\ref{tab:audit_final} of the main paper reports in the aggregate.}

\begin{figure*}[t]
    \centering
    \begin{subfigure}{1\linewidth}
        \includegraphics[width=\linewidth]{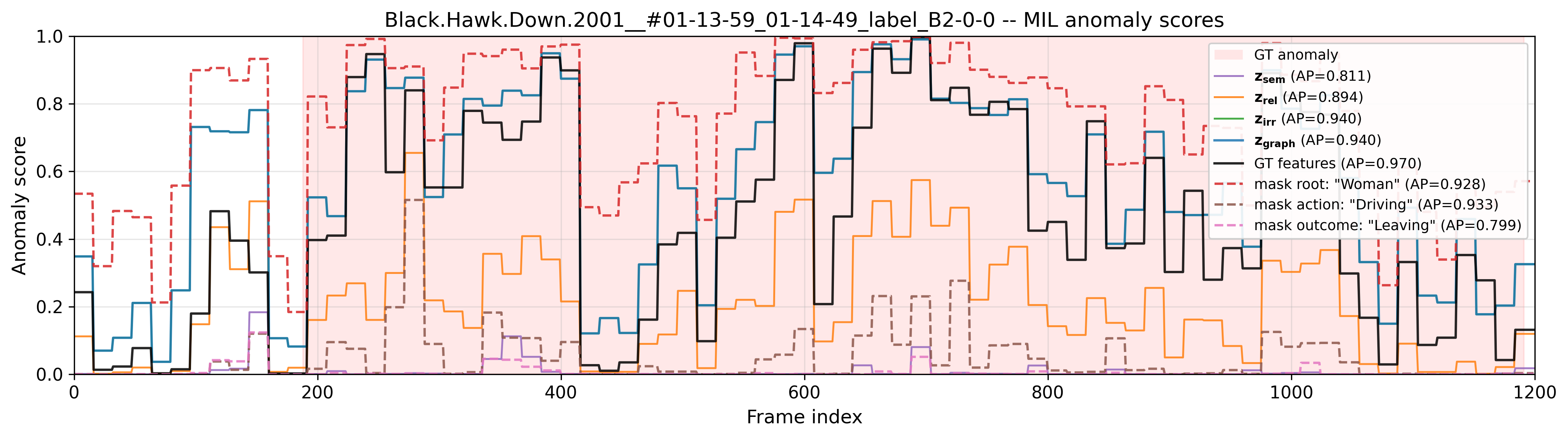}
        \caption{Black Hawk Down (Scene 1)}
        \label{fig:qual_bhd1}
    \end{subfigure}
    
    \vspace{1em}
    
    \begin{subfigure}{1\linewidth}
        \includegraphics[width=\linewidth]{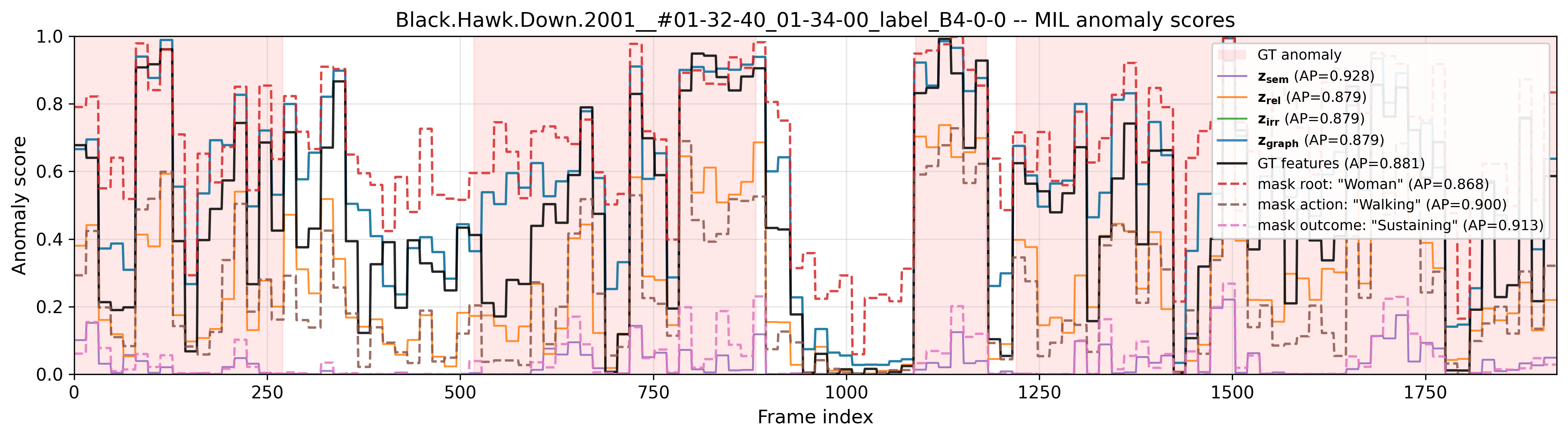}
        \caption{Black Hawk Down (Scene 2)}
        \label{fig:qual_bhd2}
    \end{subfigure}
    \caption{MIL anomaly-score traces for two \textit{Black Hawk Down} scenes under the graph masks and under masking of single named nodes (per-video AP in the legends). The same node name moves the trace differently in the two scenes; see the text.}
    \label{fig:qual_bhd}
\end{figure*}

{In Figure~\ref{fig:qual_bhd1}, masking the root node \textit{Woman} lowers the per-video AP by 0.042 against the ground-truth features (0.970 to 0.928) and by 0.012 against the full-graph reconstruction (0.940), and masking the action node \textit{Driving} behaves alike (0.933); masking the outcome node \textit{Leaving} lowers it by 0.171 (to 0.799) and pushes every score below $0.1$, a larger change than the whole relevant mask on this video (0.894). The traces also show what the AP does not. Under \textit{Woman} the scores outside the annotated window rise (more false positives) while under \textit{Driving} they fall throughout. In Figure~\ref{fig:qual_bhd2} the three masked nodes give 0.868 (\textit{Woman}), 0.900 (\textit{Walking}) and 0.913 (\textit{Sustaining}) against 0.881 for the ground-truth features and 0.879 for $\mathbf{z_{graph}}$, so two of the three single-node masks \emph{raise} the AP while lowering most scores, and semantic masking raises it further (0.928). On this video the measurement is dominated by how the masks redistribute scores across the annotated window rather than by their level.}

\begin{figure*}[t]
    \centering
    \includegraphics[width=1\linewidth]{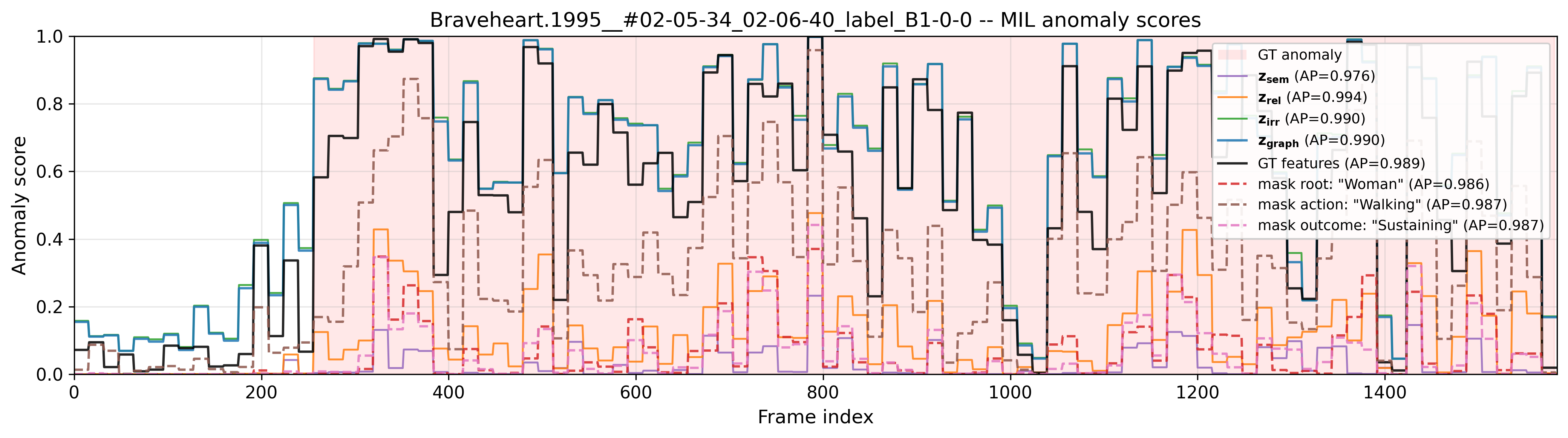}
    \caption{MIL anomaly-score traces for a \textit{Braveheart} scene. The annotated window covers most of the clip, so the single-node masks lower the scores while leaving the per-video AP within 0.004 of the ground-truth features (legend).}
    \label{fig:qual_braveheart}
\end{figure*}

{Figure~\ref{fig:qual_braveheart} is a near-saturated case, in which the ground-truth features score 0.989 AP, the full-graph reconstruction 0.990, and every single-node mask lands within 0.004 of them (\textit{Woman} 0.986, \textit{Walking} 0.987, \textit{Sustaining} 0.987) although the masked traces are visibly lower across the video; semantic masking lowers the AP to 0.976 while driving all scores below $0.5$, and the relevant mask raises it to 0.994. When the anomalous window covers most of the clip, as here, a mask can lower every score and leave the ranking, and hence the AP, almost unchanged. This is why the main paper reports test-set deltas and tests their alternative explanations, rather than per-video curves.}

\begin{figure*}[h!]
    \centering
    \begin{subfigure}{1\linewidth}
        \includegraphics[width=\linewidth]{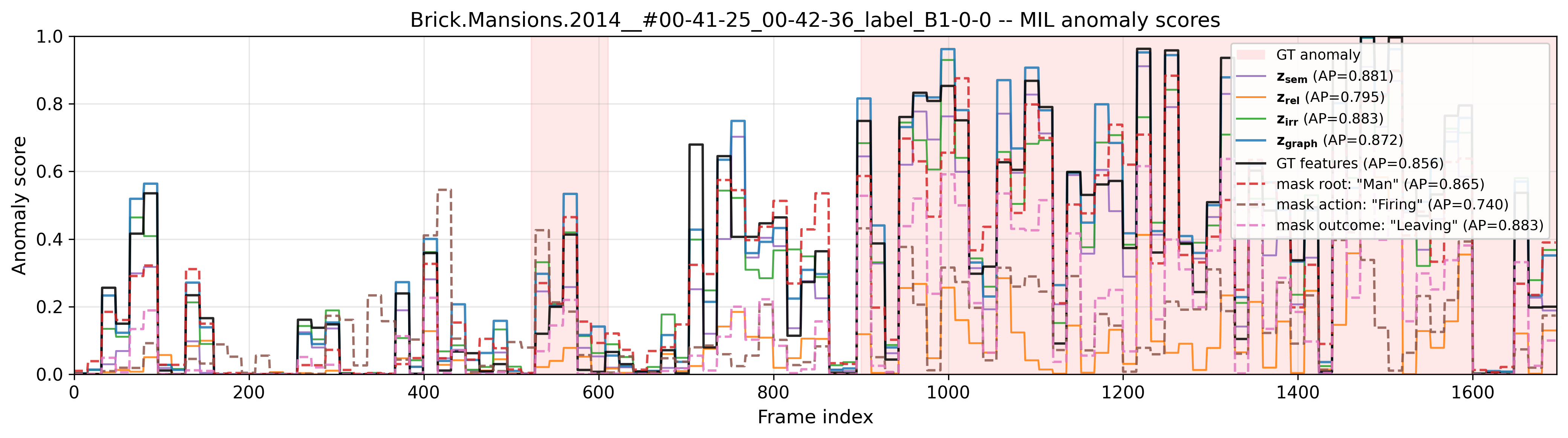}
        \caption{Brick Mansions (Scene 1)}
        \label{fig:qual_bm1}
    \end{subfigure}
    
    \vspace{1em}
    
    \begin{subfigure}{1\linewidth}
        \includegraphics[width=\linewidth]{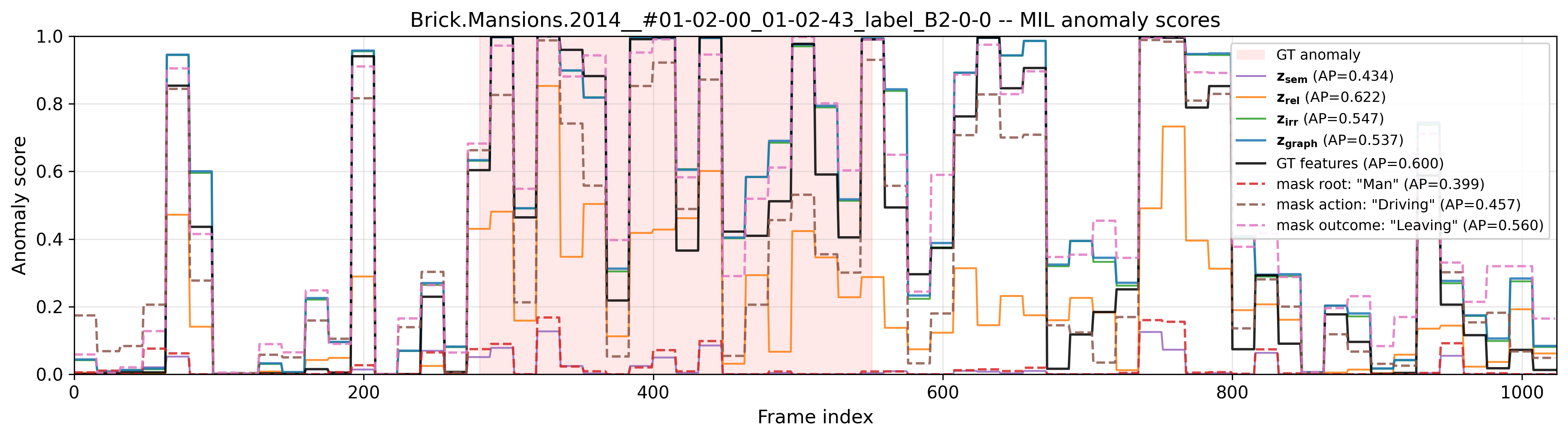}
        \caption{Brick Mansions (Scene 2)}
        \label{fig:qual_bm2}
    \end{subfigure}
    \caption{MIL anomaly-score traces for two \textit{Brick Mansions} scenes. Masking the root node \textit{Man} leaves the first scene's AP within 0.01 of the ground-truth features and lowers the second scene's from 0.600 to 0.399 (legends).}
    \label{fig:qual_bm}
\end{figure*}

{Figures~\ref{fig:qual_bm1} and~\ref{fig:qual_bm2} show the three masked nodes on two scenes of one film with very different outcomes. In the first scene masking the root node \textit{Man} changes the AP by less than 0.01 (0.865 against 0.856 for the ground-truth features and 0.872 for $\mathbf{z_{graph}}$), masking the outcome node \textit{Leaving} by 0.01 (0.883), while masking the action node \textit{Firing} lowers it to 0.740 and suppresses the scores of the normal segment (frames 630 to 900) and of the final anomalous segment while raising those between frames 200 and 600. In the second scene masking \textit{Man} lowers the AP from 0.600 to 0.399 and flattens the trace, \textit{Driving} gives 0.457 and \textit{Leaving} 0.560, and the relevant mask raises the AP to 0.622 while the full-graph reconstruction sits at 0.537. The response to a node name is therefore video-specific, and on single videos it is not monotone in what the mask removes, and the controlled, aggregate deltas of the main paper are the quantity on which the paper rests.}
\clearpage

\end{document}